\documentclass[journal, 12pt, draftclsnofoot, onecolumn]{IEEEtran}
\ifCLASSINFOpdf
\else
\fi

\usepackage{bm}
\usepackage{CJK}
\usepackage{indentfirst}
\usepackage{amsmath}
\usepackage{multirow}
\usepackage{threeparttable}
\usepackage{graphicx}
\usepackage{subfigure}
\usepackage{color}
\usepackage{booktabs}
\usepackage{epstopdf}
\usepackage{makecell}
\usepackage[linesnumbered,ruled,vlined]{algorithm2e}
\usepackage{algpseudocode}
\usepackage{amsmath}

\begin{document}
\title{IIoT-Enabled Health Monitoring for Integrated Heat Pump System Using Mixture Slow Feature Analysis}

\author{Yan~Qin,~\IEEEmembership{Member,~IEEE,}
             Wen-tai~Li,~\IEEEmembership{Member,~IEEE,}
             Chau~Yuen,~\IEEEmembership{Fellow,~IEEE,} \\
             Wayes~Tushar,~\IEEEmembership{Senior Member,~IEEE,}
             and Tapan~Kumar~Saha,~\IEEEmembership{Fellow,~IEEE}
\thanks{This work was supported in part by the National Research Foundation Singapore and administered by Building and Construction Authority (BCA)-Green Building Innovation Cluster (GBIC) Programme Office, SUTD-MIT International Design Centre, A*STAR-NTU-SUTD Joint Research Grant on Artificial Intelligence Partnership under Grant RGANS1906, and National Natural Science Foundation of China under Grant No. 61903327. Any findings, conclusions, or opinions expressed in this document are those of the authors and do not necessarily reflect the views of the sponsors. (Corresponding author: Yan Qin)}
\thanks{Y. Qin, W. Li, and C. Yuen are with the Engineering Product Development Pillar, Singapore University of Technology and Design, 8 Somapah Road, 487372 Singapore. (e-mail: yan$\_$qin@sutd.edu.sg, wentai$\_$li@sutd.edu.sg, and yuenchau@sutd.edu.sg)}
\thanks{W. Tushar and T.K. Saha are with the School of Information Technology and Electrical Engineering, University of Queensland, St. Lucia, Brisbane, Qld 4072, Australia. (e-mail: w.tushar@uq.edu.au and saha@itee.uq.edu.au)}
}

\maketitle
\begin{abstract} The sustaining evolution of sensing and advancement in communications technologies have revolutionized prognostics and health management for various electrical equipment towards data-driven ways. This revolution delivers a promising solution for the health monitoring problem of heat pump (HP) system, a vital device widely deployed in modern buildings for heating use, to timely evaluate its operation status to avoid unexpected downtime. Many HPs were practically manufactured and installed many years ago, resulting in fewer sensors available due to technology limitations and cost control at that time. It raises a dilemma to safeguard HPs at an affordable cost. We propose a hybrid scheme by integrating industrial Internet-of-Things (IIoT) and intelligent health monitoring algorithms to handle this challenge. To start with, an IIoT network is constructed to sense and store measurements. Specifically, temperature sensors are properly chosen and deployed at the inlet and outlet of the water tank to measure water temperature. Second, with temperature information, we propose an unsupervised learning algorithm named mixture slow feature analysis (MSFA) to timely evaluate the health status of the integrated HP. Characterized by frequent operation switches of different HPs due to the variable demand for hot water, various heating patterns with different heating speeds are observed. Slowness, a kind of dynamics to measure the varying speed of steady distribution, is properly considered in MSFA for both heating pattern division and health evaluation. Finally, the efficacy of the proposed method is verified through a real integrated HP with five connected HPs installed ten years ago. The experimental results show that MSFA is capable of accurately identifying health status of the system, especially failure at a preliminary stage compared to its competing algorithms. \end{abstract}


\IEEEpeerreviewmaketitle

\section{Introduction}
\IEEEPARstart{T}{he} heat pump (HP) is a high-efficiency electrical system that transfers heat from the outside building into domestic heating or cooling use \cite{Ref1}. In comparison with fossil fuel or traditional electrical-based heating solutions, HP is energy-saving and eco-friendly \cite{Ref2}, which has been widely installed from single-family houses to large commercial buildings. Recently, HPs have become increasingly popular, and the installation of HPs has experienced a fast-rising speed of 10$\%$ from 2017 to 2019 \cite{Ref3} all over the world. In practice, the healthy status of HP is a significant concern since it directly influences the efficiency of energy production, however, which may be disturbed by upset, maloperation, and equipment degradation. If no alarm is available for possible health degradation, the under-performance status of HP will continue and may even evolve into serious safety violations. Therefore, providing reliable and accurate health degradation monitoring method is urgent and crucial for HP.

Health monitoring from prognostic and health management technology \cite{Ref4}-[20] has been developed against possible faults in guaranteeing safety, reliability, and efficiency, which typically consists of two sequential parts, offline modeling and online monitoring. In the offline part, normal working patterns are learned from apriori knowledge, operation experiences, physicochemical reactions, and historical data, etc. In the online part, new information (expertise, data, etc.) is then compared with the known normal patterns to check whether normal status holds. According to the way to acquire normal patterns, current methods are broadly grouped into knowledge-based, model-based, and data-driven approaches \cite{Ref4}. Since it consumes intensive labor and time-consuming costs to derive expert knowledge and physical models, data-driven approaches are ever-increasing popular with rapid advancements in data acquisition, storage, and calculation.

Multivariate statistical process control (MSPC) has been recognized as an effective data-driven health monitoring for industrial processes and safety-critical assets over the past decades. The idea behind of MSPC is to integrate latent variables projection and statistical modeling systematically. Latent variables typically are linear combinations of original measurements, together with given specific names and meanings in different algorithms. Specifically, principal component (PC) is named in principal component analysis (PCA) to convey main variations, which plays a milestone role in MSPC when the systems are straightforward. Kresta et al. \cite{CJCE_PCA} separated the derived PCs into two independent subspaces according to the degree of process variations. On the basis of this, statistical control limit $T^2$ can be established for the system subspace containing most variations with dominant PCs, and control limit squared prediction error ($SPE$) will be developed for the residual subspace with the remaining PCs. In the same framework, variants of PCA have been reported to deal with various practical concerns. For instance, non-linear PCA was proposed for handling the variable nonlinearity \cite{NonlinearPCA}, \cite{NonlinearPCA_Li}; multi-way PCA \cite{Multiway_PCA} and step-wise PCA \cite{stepwisePCA} have been developed to fit three-dimensional data structure raised by batch processes. For HPs, Chen et al. \cite{HP_PCA1} employed PCA for performance monitoring of air-source HP. Likewise, Hu et al. \cite{HP_PCA2} applied PCA for fault detection of HP-based chiller. Further, temporal dynamics, which is widely observed in industrial systems, has been analyzed by dynamic latent variables to increase the sensitivity to incipient faults \cite{DLV}, \cite{JPC_Zhu}. For conciseness, the interested readers are encouraged to find more details and useful information in \cite{Survey1}, \cite{Survey2}. The methods mentioned above require data following Gaussian distribution, in which homogeneous data distribution should be ensured for constructing control limits. To address the non-Gaussian situation, Fetial et al. \cite{GMM_Monitoring} adopted the Gaussian mixture model (GMM) for clustering data from the same distribution, and PCA-based monitoring models were developed in each local model with Gaussian distribution.

Over the past decades, the modern industries have become increasingly complex and large-scale with automation devices to save cost and improve production efficiency. One benefit is that multiple products could be produced or different product proportions could be online adjusted in the same system. For instance, various plastics products can be produced using the same injection molding machine by just changing materials or replacing molds \cite{AICHE_QIN}. This normal shifts of operating conditions raise new challenges for approaches developed at a specific condition since they mainly exploit static information for monitoring purpose. An example is given here to better understand the static information and dynamic information. For instance, to accurately track the trajectory of a moving object in plain view, we not only need the instant position and coordinates (static information), but also its velocity and angle steering (dynamic information). Recently, the superiority of slow feature analysis (SFA) in analyzing slow-varying dynamics has been exploited for health motioning under varying operating conditions. SFA is originally designed for identifying slowly varying features and the associated dynamic information from rapidly varying measurements \cite{SFA2002}.  With a monitoring model developed at a starting operating condition, new conditions could be distinguished not only from each other, but also from fault cases when static information is beyond the control limits, while dynamic information is normal. Although this kind of methods \cite{RecursiveSFA}, \cite{SFABatch} have achieved remarkable success in handling varying operating conditions, they are suitable for the system in which the status shifts are caused by static deviation rather than dynamics deviation, such as catalyst deactivation, cleaning, and etc.

Practically, multiple HPs will be combined to construct a more powerful integrated HP (IHP) system along with a large water tank since the tremendous demand for hot water in commercial buildings, hospitals, schools, etc. It is reasonable to turn on more HPs to accelerate heating when the hot water demand is high and turn off some HPs to save energy when the hot water demand is low. Consequently, IHP faces more complex operating conditions than industrial processes due to frequent on/off of pumps with variable hot water demand. Various operating schedules with specific heating dynamics are observed because of the different combinations of running HPs. Correspondingly, data characteristics of IHP presents heterogeneous distributions not only in static distribution but also in dynamic space. Although the remarkable ability of SFA makes it preferred as the monitoring model, it is still challenging to develop a health monitoring model for IHP with varying static and dynamic behaviors. Here, we summarize problems of health monitoring for IHP when SFA is employed:
{\begin{itemize}
\item Historical data from IHP are continuous time-series experiencing various patterns randomly in both static and dynamic spaces. No aprior knowledge is available to label how many patterns and when a specific pattern starts and ends. We argue that an unsupervised sequential clustering that groups similar time-series into patterns will improve process insights and benefit the following monitoring performance. Although GMM enables clustering samples simultaneously, the samples classified in each group may be discontinuous in the time dimension. However, this is not the actual case in practical scenario, and it is more reasonable to group adjacent samples with similar data distribution and temporal correlation into the same cluster.
\item The prerequisite of SFA-based health monitoring does not hold since the dynamics of each pattern may be different. We need to design a monitoring strategy for an IPH system that can distinguish each pattern with specific static and dynamic information from real faults, so as to enhance the monitoring sensitivity and reduce the false alarm rate.
\item The HPs have served many years and may not possess necessary sensors, which may be caused by the behindhand sensor technology and high-priced cost at that time. From the point of view of practical implementation, it is worthy of designing an economical and affordable way to facilitate the health monitoring algorithm.
\end{itemize}}

With the increasing evolution of sensing and wireless network technologies, we argue that it is possible to use industrial Internet-of-Things (IIoT) technology \cite{Ref10}, \cite{Ref11} to collect valuable data information for analysis rather than replacing or upgrading HPs. It is a promising solution to deploy IIoT sensors that have a close relationship with the health status of HP to record data. Furthermore, to address the technical challenges concerning health monitoring for IHP, we propose a novel framework by integrating IIoT technology and intelligent algorithm to develop an IIoT-enabled health monitoring strategy for IHP without built-in sensors. First, the proposed framework deploys temperature sensors at the inlet and outlet of the water tank to sense water temperature. Next, an unsupervised learning-based health monitoring algorithm named mixture slow feature analysis (MSFA) is proposed to serve as health evaluation for IHP. MSFA consists of temporal pattern clustering and fine-grained status detection for both off-line modeling and online evaluation. Here, the temporal sequential clustering-based GMM is capable of identifying patterns with different data distributions in both steady and dynamical spaces. Then in each pattern, a hierarchical evaluation system is constructed with upper layer steady process variations and the lower layer dynamic information concerning process variations. With global probability indices developed by weighting each pattern, a set of rules are given to accurate identify the fault. To verify the efficacy of the proposed framework, we study a practical IHP with five connected HPs for heating water and carefully compare it with the typical MSPC approaches. The main contributions of the proposed method are summarized below:

(1) A novel framework is put forward for the first time to benefit the existing IHP system to evaluate their health status by integrating IIoT technology and machine learning.

(2) We introduce temporal correlations into GMM-based sequential clustering, ensuring the classification results are sequential.

(3) A health monitoring strategy is proposed to overcome the inability of SFA for the IHP system under frequent and repeat varying working conditions, in which both static and dynamics are different.

\begin{figure}
\centering
\includegraphics[scale=0.5]{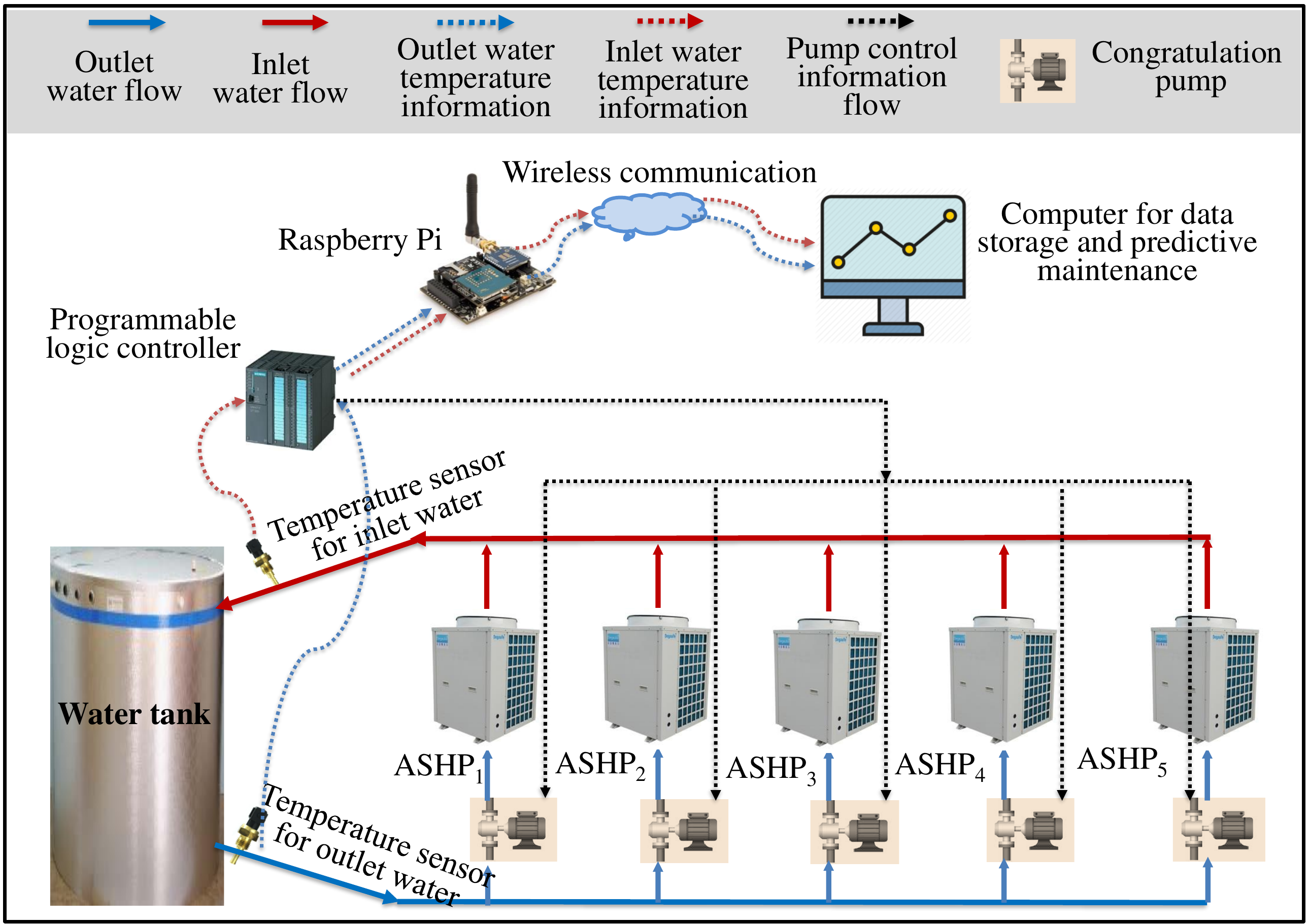}\\
\caption{The basic structure of the employed IHP and sensor locations.}
\label{MyFig1}
\vspace{-0.5cm}
\end{figure}

The organization of this paper is arranged as follows: Section II exhibits the basic structure of IIoT-enabled IHP and the data characteristics of multiple patterns. The details of the proposed MSFA are shown in Section III. With a practical application, Section IV illustrates the efficacy of the proposed method. The conclusions are drawn in Section V.

\section{IIoT-Enabled Integrated Heat Pump System Setup}
The employed IHP has already been used for ten years at a hospital building located in Singapore. The IHP considered in this paper consists of a large tank for hot water storage and five paralleled air-sourced HP (ASHP) units, as shown in Fig. 1. Each unit consists of an ASHP and its corresponding congratulation pump. When the IHP starts, ASHPs work in heating mode to absorb heat from warm air. Meanwhile, water flow is pumped from the middle of the water tank to the inlet of ASHPs through a corresponding congratulation pump. With produced heat, water is heated when it passes through ASHPs. At last, heated water returns to the water tank for daily use. By cycling the heating procedures, water temperature is increasingly raised until it reaches its set-point.

The IIoT-enabled system is set up by installing two temperature sensors at the inlet and outlet of the water tank, as shown in Fig. 1. The reason that the temperature sensor is preferred as follows: water temperature is mostly related to the health status of the concerned ASHPs, which determines an important metrics coefficient of performances of ASHP. The programmable logic controller (PLC) is used to receive water temperature from sensors, and corresponding control signals are acted on pumps. Raspberry Pi is developed to serve as a local gateway to retrieve the inlet and outlet water temperatures, and heat pump on/off timing from the PLC and send the readings to a cloud server over a cellular network. The sampling rate of water temperatures is 1 minute. The details on the hardware responsible for HP specifications, sensor type, and data communication are listed in Table I.

\begin{figure}
\centering
\includegraphics[scale=0.5]{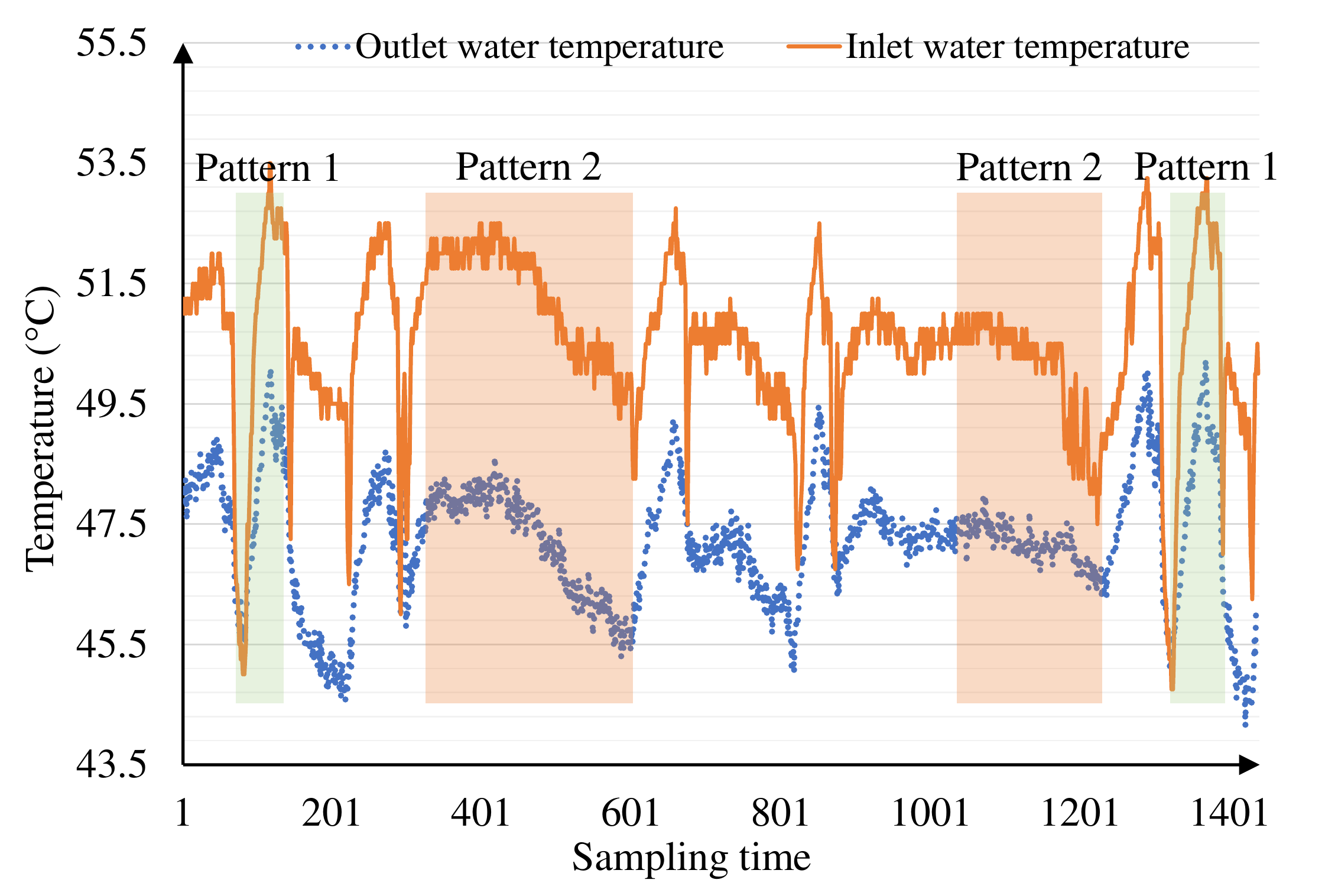}\\
\caption{The varying trend of measured inlet water and outlet water temperature collected from the test-bed on 1 April 2018.}
\label{MyFig2}
\vspace{-0.4cm}
\end{figure}

\begin{table}
\scriptsize
\renewcommand{\arraystretch}{1.2}
\setlength{\abovecaptionskip}{-0.3cm}
\setlength{\belowcaptionskip}{-0.3cm}
\caption{Hardware configuration of the IoT-enabled IHP system.}
\label{table_example}
\begin{center}
\begin{threeparttable}
\begin{tabular}{p{4cm} p{3cm} p{6cm}}
\toprule
Name & Type & Description \\
\hline
ASHP 1 to 5 & DGL-50C (Degaulle) & Installation in 2010, and the heating capacity of each one is 52.1(kW)\\
\hline
Inlet water temperature sensor    & Pt100 & Accuracy is 0.1$^\circ$C $\pm$ 0.17$\%$, and the sampling rate is 1 minute\\
\hline
Outlet water temperature sensor & Pt100 & Accuracy is 0.1$^\circ$C  $\pm$ 0.17$\%$, and the sampling rate is 1 minute \\
\hline
Water tank  &  & 12000 (liter)\\
\bottomrule
\end{tabular}
\end{threeparttable}
\end{center}
\vspace{-0.6cm}
\end{table}

In the setup, the PLC is programmed such that if the temperature of inlet water is below 46$^\circ$C, ASHPs will be activated to heat water, and they are stopped if the corresponding water temperature higher than 52$^\circ$C. To prolong the life of the ASHPs and avoid overusing them, ASHPs will be turned on in an alternating manner. This algorithm is implemented to balance out the workload of individual ASHP according to the hot water requirements of the building. More information about the control schedules can be found in \cite{Ref12}.

Fig. 2 exhibits the collected data regarding two water temperatures on April 1 in 2018 for a better understanding of the data characteristics. It is observed that the measurements are time-varying at a rapidly changing speed due to the frequent adjustments of ASHPs. Furthermore, similar patterns are repeatedly appeared, such as the green shadow and the pink shadow marked in Fig. 2, revealing the existence of multiple data patterns. Although the combinations of ASHPs are various, the heating ability may be similar for some of them. Similar patterns need to be merged, and the dissimilar ones should be well classified for better performance monitoring. Here, we argue that three unique properties need careful consideration for pattern classification and health evaluation in this article:

(1) Since the water temperature continually varies, it is reasonable to classify adjacent samples into the same class. That is, the classification results are sequential with consideration of temporal correlations.

\begin{figure*}
\centering
\includegraphics[scale=0.36]{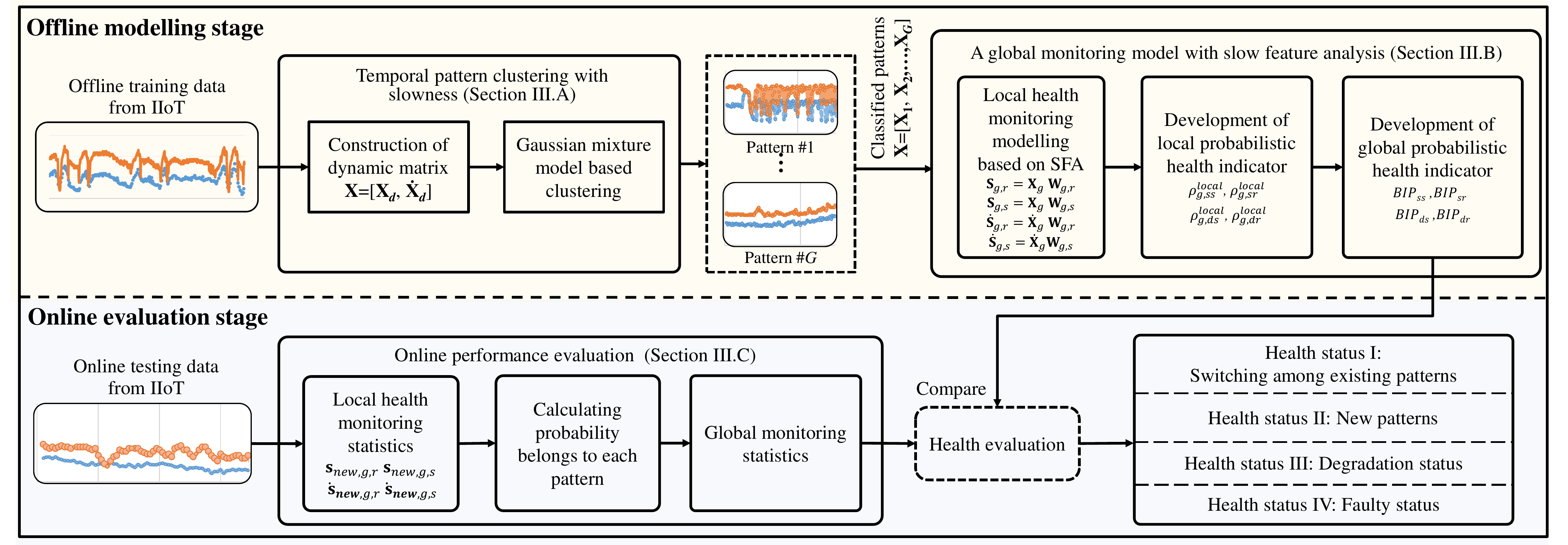}\\
\caption{The basic structure of the proposed MSFA algorithm.}
\label{MyFig4}
\vspace{-0.7cm}
\end{figure*}

(2) Due to the frequent on/off of the pump, it requires the ability to handle more complex situations for IHP data. Several concepts used in the following text are explained from the perspective of data distribution here for better understanding. New pattern: the steady data distribution of incoming data does not belong to any one of existing patterns, but its dynamics keeps in consistency with that of the existing patterns. This is may be caused by the scheduled maintenance, such as replacing or filling with free-flowing lubricating oil. Faulty status: the behavior of both steady distribution and dynamics deviate from their normal region.

(3) It is beneficial to consider varying speed of variables, contributing to well distinguish patterns with the same steady distribution but different varying speeds. We use slowness, which is calculated as the difference of a variable, to present this kind of process dynamics.

\section{Mixture slow feature analysis}
In this section, a temporal clustering and health monitoring method named mixture slow feature analysis (MSFA) is implemented as shown in Fig. 3, which includes three parts. The first part details the sequential pattern clustering with considerations of temporal correlation and slowness. Second, a global monitoring model with classified patterns is put forward by integrating steady and dynamic information in each pattern. At last, the online procedure is conducted to provide reliable monitoring results about system health status.

\subsection{Temporal Pattern Clustering with Slowness}

Multiple data distributions are observed in IHP due to variously operation schedules. To cluster similar patterns, two issues deserve careful consideration rather than directly adopting GMM. The first one is how to ensure the sequential clustering characteristic because of the inherent temporal correlations and avoid adjacent samples being assigned into distinct clusters. The second one is how to separate patterns that are similar in steady distribution but different in dynamics.

To achieve these purposes during pattern clustering, lagged data vector $ \bm{\textbf x}_d(k)=[{\textbf x}(k),{\textbf x}(k-1),..., {\textbf x}(k-h+1)]$ is first constructed to replace the original sample ${\textbf x}(k)$ at sampling time $k$, in which $h$ is the value of time lag and can be determined by sampling self-correlation analysis. Through data arrangement at each time, a dynamic data matrix $\bm{\textbf X}_d$ is obtained as $\bm{\textbf X}_d=[\bm{\textbf x}_d(h), \bm{\textbf x}_d(h+1),..., \bm{\textbf x}_d(K)]$, where $K$ is the total number of samples. In this way, temporal extension of original data matrix contributes to sequential clustering. Further, to consider dynamics of extended matrix $\bm{\textbf X}_d$, its first-order difference information is calculated as $\dot{\textbf X}_d=[\dot{\textbf x}_d(h), \dot{\textbf x}_d(h+1),..., \dot{\textbf x}_d(K-1)]$, in which $\dot{\textbf x}_d(h)=x_d(h+1)-x_d(h)$. Merging ${\textbf X}_d$ and $\dot{\textbf X}_d$ by keeping sample-wise unchanged, a new matrix $\textbf X$=$[{\textbf X}_d, \dot{\textbf X}_d]$ taking the varying speed of measurements into account is constructed for the following clustering analysis.

GMM aims at separating and classifying mixed data with different data distributions into several local centers, each of which follows a specific Gaussian distribution \cite{Ref13}. The strengths of GMM have received a flurry of attention in multimodal optimization \cite{GMMoptimization}, source separation \cite{GMMsource}, etc.
For variables $\mathbf{x} \in \mathbf{R}^{J}$ in data matrix $\bm{\textbf X}$, its distribution can be approximated by weighting $G$ Gaussian components,
\begin{equation}
\begin{array}{c}{p(\mathbf{x})=\sum_{g=1}^{G} \pi_{g} N(\mathbf{x} | \boldsymbol{\mu}_{g}, \mathbf{\Sigma}_{g})} \\ [1mm]
{ { s.t. } \sum_{g=1}^{G} \pi_{g}=1}
\end{array}
\end{equation}
where $N(\cdot)$ follows a Gaussian distribution with mean $\bm \mu_g$ and covariance matrix $\bm \Sigma_{g}$ of the $g^{th}$ Gaussian component, and $J$ is the number of variables in $\textbf X$. Besides, the constraint normalizes the significance of each Gaussian component with non-negative weights $\pi_{g}$.

Assuming that $\textbf x$ belongs to the $g^{th}$ Gaussian component, its Gaussian density function is defined as follows,
\begin{equation}
\begin{aligned} p(\mathbf{x} | \mathbf{x} \in g) &=N(\mathbf{x} |(\boldsymbol{\mu}_{g}, \mathbf{\Sigma}_{g})) \\ &=\frac{\exp \{-\frac{1}{2}(\mathbf{x}-\boldsymbol{\mu}_{g})^{T} \mathbf{\Sigma}_{g}^{-1}(\mathbf{x}-\boldsymbol{\mu}_{g})\}}{(2 \pi)^{J / 2}|\mathbf{\Sigma}_{g}|^{1 / 2}}
\end{aligned}
\end{equation}

For all Gaussian components and their associated coefficients, $\mathbf \Theta = \{ {\bm \mu_1, \bm \Sigma_1, \pi_1, \bm \mu_2, \bm \Sigma_2, \pi_2,..., \bm \mu_G, \bm \Sigma_G, \pi_G} \}$ need to be estimated to describe the data distribution of $\textbf x$. Typically, expectation maximization (EM) \cite{Ref14} is used to solve this problem by maximizing log-likelihood probability below,
\begin{equation}
\hat{\mathbf \Theta}=\arg \max _{\theta} \log p(\mathbf{X} | \mathbf \Theta)
\end{equation}
where,
\begin{equation}
\begin{aligned}
\log p(\mathbf{X} | \mathbf{\Theta}) & = \log \prod_{k=1}^{K} p(\mathbf{x}_{k} | \mathbf{\Theta}) \\
& = \sum_{i=1}^{K} \log \sum_{g=1}^{{G}} \pi_{g} p(\mathbf{x}_{k} | (\bm \mu_g, \bm \Sigma_g))
\end{aligned}
\end{equation}

We differentiate Eq. (4) with respect to $\bm \mu_g$ and set the expression equal to zero as below,
\begin{equation}
\sum_{k=1}^{K} \frac{\pi_{g} N(\bm x_{k} | \bm \mu_{g}, \bm \Sigma_{g})}{\sum_{g=1}^{G} \pi_{g} N(\bm x_{k} | \bm \mu_{g}, \bm \Sigma_{g})} \frac{(\bm x_{k}-\bm \mu_{g})}{\bm \Sigma_{g}}=0
\label{5}
\end{equation}

Step 1. Initialize {$\pi_{g}$, $\bm \mu_g$, $\bm \Sigma_{g}$} and evaluate the Eq. (4) with these parameters.

Step 2. Introduce the latent variable $Z_g$, and calculate the posterior distribution of given the observations $\mathbf x_k$ as below,
\begin{equation}
\begin{aligned}
P(Z_{g} = g \mid \mathbf x_{k}) & = \frac{P(\mathbf x_{k} \mid Z_{g} = g) P(Z_{k}=g)}{P(\mathbf x_{k})} \\
& =\frac{\pi_{k} N(\bm \mu_{g}, \bm \Sigma_{g})}{\sum_{g=1}^{G} \pi_{g} N(\bm \mu_{g}, \bm \Sigma_{g})} =\gamma_{k}(g)
\end{aligned}
\end{equation}

Step 3. Solve $\bm \mu_{g}$ by derivative respect to $\bm \mu_{g}$ through combining Eqs. (5) and (6), which is given below,
\begin{equation}
\begin{aligned}
\hat{\bm \mu}_{g}=\frac{\sum_{k=1}^{K} \gamma_{k}(g) \mathbf x_{k}}{\sum_{k=1}^{K} \gamma_{k}(g)}=\frac{1}{N_g} \sum_{k=1}^{K} \gamma_{k}(g) \mathbf x_{k}
\end{aligned}
\end{equation}
where $K_{g}=\sum_{k=1}^{K} \gamma_{k}(g)$.

Step 4. Calculate $\hat{\bm \Sigma}_{g}$ and $\hat{\bm \pi}_{g}$ in a similar way to $\bm \mu_{g}$, which are given below,
\begin{equation}
\begin{aligned}
\hat{\bm \Sigma}_{g} &=\frac{1}{K_{g}} \sum_{k=1}^{K} \bm \gamma_{k}(g)(x_{k}-\bm \mu_{g})^{2} \\
\hat{\pi}_{g} &=\frac{K_{g}}{K}
\end{aligned}
\end{equation}

Step 5. Evaluate the log-likelihood with the estimated new parameters. The iterative procedure could be stopped if the log-likelihood value has reduced by less than a small preset value. Otherwise, go back to Step 2 for the next iteration.

With $\hat {\mathbf \Theta}$, samples in $\textbf X$ are assigned into the corresponding Gaussian component, where the maximum probability of a certain sample is observed. Let the divided subgroups of $\textbf X$ be $\{ \textbf X_1$, $\textbf X_2$, ..., $\textbf X_g$, ..., $\textbf X_G\}$, each of which corresponds to the derived Gaussian distribution. It should be noted that the initial value of $G$ can be determined according to all possible combinations of ASHPs, and then clustered patterns with both similar steady and dynamic information will be merged.

\subsection{A Global Monitoring Model with Slow Feature Analysis}
As a method to find feature spaces, where variations of latent variables vary as slowly as possible, SFA is good at recognizing slow-varying dynamics \cite{Ref15}. Inspired by this, SFA is performed in each pattern to extract latent variables presenting different slowness.

\subsubsection{Local health monitoring modeling based on SFA}
Given a feature space spanned by a series of linear mapping functions $\{ \bm \omega_1(\cdot), \bm \omega_2(\cdot),...,\bm \omega_q(\cdot) \}$, data matrix $\textbf X_g(K_g{\times}J)$ in the $g^{th}$ pattern is first normalized to zero mean and unit variance using the estimated $\bm {\mu}_g$ and $\bm {\Sigma}_g$ in Eq. (4), in which $K_g$ is the number of samples in the $g^{th}$ mode. Then $\widetilde {\textbf X}_g$, which is lagged measurement from normalized $\textbf X_g$, is projected on weight vector $\bm \omega_i$ to get the corresponding slow features, i.e., ${\textbf s}_i={\widetilde {\textbf X}_g} {\bm \omega}_i$. The finding of slow features is equal to search for a series of weighting vectors $\bm \omega$  by minimizing the variance of temporal difference below,
\begin{equation}
\begin{array}{l}
{\Delta(\mathbf{s}_{i}):=\langle\mathbf{\dot s}_{i}^{2}\rangle} \\ [1mm]
{{s.t.} \langle\mathbf{s}_{i}\rangle= 0} ; {\langle\mathbf{s}_{i}^{2}\rangle= 1}; {\langle\mathbf{s}_{i} \mathbf{s}_{j}\rangle= 0 \forall j<i}\end{array}
\end{equation}
where $\dot {\textbf s}_i$ is the first-order temporal difference of $\textbf {s}_i$. The first two constraints in Eq. (9) ensure the mean and variance of $\textbf {s}_i$ are zero and unit, respectively. Besides, redundant information between slow features is avoided by keeping the independence of each slow feature in the third constraint.

The weight vector $\bm \omega$ is calculated by solving the generalized eigenvalue problem as follows,
\begin{equation}
{\bm \Sigma}_{g} {\bm \omega}={\lambda} {\bm \Xi_{g}} {\bm \omega}
\vspace{-0.1cm}
\end{equation}
where ${\bm \Sigma}_g$ is the covariance of $\widetilde {\textbf X}_g$; ${\bm \Xi}_g$ is the covariance of $\dot {\widetilde {\textbf X}}_g$, in which $\dot {\widetilde {\textbf X}}_g$ is the first-order temporal difference of $\widetilde {\textbf X}_g$; $\lambda$ is the eigenvalue reflecting the slowness of slow feature, which is sorted in descending order.

Denote the collection of weight vectors obtained in Eq. (10) as matrix $\textbf {W}_g = [\bm \omega_{g,1}, \bm \omega_{g,2},...,\bm \omega_{g,J}]$. Latent variables $\textbf {S}_g = [\textbf s_{g,1}, \textbf s_{g,2},...,\textbf s_{g,J}]$ located in the $g^{th}$ pattern are computed by projecting $\widetilde {\textbf X}_g$ on $\textbf W_g$, in which ${\textbf s}_{g,i}={\widetilde {\textbf X}_g} {\bm \omega}_{g,i}$. According to the values of slowness, latent variables in $\textbf S_g$ are classified into two groups. The first group is slow feature that presents slow-varying process variation, and the second group consists of latent variables that  vary fast. The number of slow features can be determined by the criterion of slowness, which is sequentially reduced in Eq. (10) \cite{Ref15}. Assuming the number of retained slow features is $P_g$, the corresponding weighting matrix is denoted as $\textbf {W}_{g,s} = [\bm \omega_{g,1}, \bm \omega_{g,2},...,\bm \omega_{g,P_g}]$, which is the first $P_g$ columns of $\textbf W_g$. Correspondingly, the latent variables in the residual subspace are denoted as $\textbf S_{g,r}$ , in which ${\textbf S}_{g,r}={\widetilde {\textbf X}_g} {\mathbf W}_{g,r}$ and $\mathbf {W}_{g,r} = [\bm \omega_{g,P_g+1}, \bm \omega_{g,P_g+2},...,\bm \omega_{g,J}]$. Furthermore, the dynamic matrix $\dot {\textbf S}_{g,s}$ and $\dot {\textbf S}_{g,r}$ of slow-varying subspace and its residual subspace are computed below,
\begin{equation}
\begin{array}{l}
{\dot{\mathbf{S}}_{g,s}=\dot {\widetilde {\textbf X}}_g \mathbf{W}_{g,s}} \\ [1mm]
{\dot {\mathbf{S}}_{g,r}=\dot {\widetilde {\textbf X}}_g \mathbf{W}_{g,r}}\end{array}
\end{equation}

In summary, four specific subspaces are derived in each pattern, which are steady subspaces $\textbf S_{g,s}$ and $\textbf S_{g,r}$, and their dynamic subspaces $\dot {\textbf S}_{g,s}$ and $\dot {\textbf S}_{g,r}$.

\begin{table*}[!ht]
\tiny	
\renewcommand{\arraystretch}{1.3}
\setlength{\abovecaptionskip}{-0.4cm}
\setlength{\belowcaptionskip}{-2cm}
\caption{Rules for evaluation the health status for IHP with the defined global monitoring indices.}
\label{Table1}
\begin{center}
\begin{threeparttable}
\begin{tabular}{p{1.5cm} p{2.6cm} p{2.6cm} p{2.6cm} p{2.6cm} p{4cm}}
\toprule
\multicolumn{1}{c}{\multirow{2}{*}{\textbf{Health status}}} & \multicolumn{2}{c}{\textbf{Static indices}} & \multicolumn{2}{c}{\textbf{Dynamic indices}} & \multicolumn{1}{c}{\multirow{2}{*}{\textbf{Descriptions}}} \\
\cline{2-5}
& \multicolumn{1}{c}{$BIP_{ss}$} & \multicolumn{1}{c}{$BIP_{sr}$} & \multicolumn{1}{c}{$BIP_{ds}$} & \multicolumn{1}{c}{$BIP_{dr}$} & \\
\hline
\multicolumn{1}{c}{\multirow{3}{*}{Normal switching (I)}} & $BIP_{ss}(k-2:k)<1-\alpha$ & $BIP_{sr}(k-2:k)<1-\alpha$ & $BIP_{ds}(k)>1-\alpha$ & $BIP_{dr}(k-2:k)<1-\alpha$ & \multirow{2}{*}{\makecell[c]{Sporadic dynamic abnormality is observed with
\\ index $BIP_{ds}$ or $BIP_{dr}$, 
\\ meanwhile both steady indices are normal.}} \\
\cline{2-5}
& $BIP_{ss}(k-2:k)<1-\alpha$ & $BIP_{sr}(k-2:k)<1-\alpha$ & $BIP_{ds}(k-2:k)<1-\alpha$ & $BIP_{dr}(k)>1-\alpha$ & \\
\cline{2-5}
& $BIP_{ss}(k-2:k)<1-\alpha$ & $BIP_{sr}(k-2:k)<1-\alpha$ & $BIP_{ds}(k)>1-\alpha$ & $BIP_{dr}(k)>1-\alpha$ & \\
\hline
\multicolumn{1}{c}{\multirow{3}{*}{New patterns (II)}} & $BIP_{ss}(k-2:k)>1-\alpha$ & $BIP_{sr}(k-2:k)<1-\alpha$ & $BIP_{ds}(k-2:k)<1-\alpha$ & $BIP_{dr}(k-2:k)<1-\alpha$ & \multirow{2}{*}{\makecell[c]{The dynamics of the IHP is still keeping normal. \\
The deviation of steady distribution may be caused by \\
the natural equipment degradation or manual maintenance.}} \\
\cline{2-5}
& $BIP_{ss}(k-2:k)<1-\alpha$ & $BIP_{sr}(k-2:k)>1-\alpha$ & $BIP_{ds}(k-2:k)<1-\alpha$ & $BIP_{dr}(k-2:k)<1-\alpha$ & \\
\cline{2-5}
& $BIP_{ss}(k-2:k)>1-\alpha$ & $BIP_{sr}(k-2:k)>1-\alpha$ & $BIP_{ds}(k-2:k)<1-\alpha$ & $BIP_{dr}(k-2:k)<1-\alpha$ & \\
\hline
\multicolumn{1}{c}{\multirow{3}{*}{Degradation (III)}} & $BIP_{ss}(k-2:k)<1-\alpha$ & $BIP_{sr}(k-2:k)<1-\alpha$ & $BIP_{ds}(k-2:k)>1-\alpha$ & $BIP_{dr}(k-2:k)<1-\alpha$ & \multirow{2}{*}{\makecell[c]{The dynamics of IHP has been changed owing to upset \\
or the degradation of IHP, but the fault is not serious \\
enough to influence its steady behavior.}} \\
\cline{2-5}
& $BIP_{ss}(k-2:k)<1-\alpha$ & $BIP_{sr}(k-2:k)<1-\alpha$ & $BIP_{ds}(k-2:k)<1-\alpha$ & $BIP_{dr}(k-2:k)>1-\alpha$ & \\
\cline{2-5}
& $BIP_{ss}(k-2:k)<1-\alpha$ & $BIP_{sr}(k-2:k)<1-\alpha$ & $BIP_{ds}(k-2:k)>1-\alpha$ & $BIP_{dr}(k-2:k)>1-\alpha$ & \\
\hline
\multicolumn{1}{c}{\multirow{3}{*}{Faulty status (IV)}} & $BIP_{ss}(k-2:k)>1-\alpha$ & $BIP_{sr}(k-2:k)<1-\alpha$ & $BIP_{ds}(k-2:k)>1-\alpha$ & $BIP_{dr}(k-2:k)<1-\alpha$ & \multirow{2}{*}{\makecell[c]{This means that IHP has already stepped into \\
a faulty status, as obvious abnormality can \\
be observed in steady distribution. }} \\
\cline{2-5}
& $BIP_{ss}(k-2:k)<1-\alpha$ & $BIP_{sr}(k-2:k)>1-\alpha$ & $BIP_{ds}(k-2:k)<1-\alpha$ & $BIP_{dr}(k-2:k)>1-\alpha$ & \\
\cline{2-5}
& $BIP_{ss}(k-2:k)>1-\alpha$ & $BIP_{sr}(k-2:k)>1-\alpha$ & $BIP_{ds}(k-2:k)>1-\alpha$ & $BIP_{dr}(k-2:k)>1-\alpha$ & \\
\bottomrule
\end{tabular}
\end{threeparttable}
\end{center}
\vspace{-0.7cm}
\end{table*}

To evaluate the normal region of the $g^{th}$ pattern with respect to steady and dynamic variations, four monitoring statistics, i.e., indicators, are designed corresponding to the derived four subspaces in the last subsection, which can be classified into two groups. The first group aims at steady variations $\textbf S_{g,s}$ and $\textbf S_{g,r}$. And the second group aims at monitoring dynamic variations $\dot {\textbf S}_{g,s}$ and $\dot {\textbf S}_{g,r}$.

Since slow features $\textbf S_{g,s}$ is unit variance, Hoteling's $T^2$ statistics is employed to derive control limit (alarm threshold), which is approximated by an $\chi^{2}$ distribution \cite{Ref16} below,
\begin{equation}
T_{g,s}^{2}=\mathbf{S}_{g,s}^{T} \mathbf{S}_{g,s} \sim \xi_{g,s} \chi_{h_{g,s}, \alpha_{g,s}}^{2}
\end{equation}
where $\xi_{g,s}=v_{g,s}/2m_{g,s}$ and $h_{g,s}=2\left(m_{g,s}\right)^{2}/v_{g,s}$, $m_{g,s}$ is the average of $\operatorname{diag}\left(\mathbf{S}_{g, s}^{\mathrm{T}} \mathbf{S}_{g, s}\right)$, $v_{g,s}$ is the corresponding variance, and $\alpha_{g,s}$ is the significant level (here is 0.01) to derive the 99$\%$ confidence limit.

Similarly, the monitoring statistic $T_{g,r}^2$  following a $\chi^{2}$ distribution is designed for the steady residual subspace, which is given below,
\begin{equation}
T_{g,r}^{2}=\mathbf{S}_{g,r}^{T} \mathbf{S}_{g,r} \sim \xi_{g,r} \chi_{h_{g,r}, \alpha_{g,r}}^{2}
\end{equation}
where $\xi_{g,r}=v_{g,r}/2 m_{g,r}$ and $h_{g,r}=2\left(m_{g,r}\right)^{2}/v_{g,r}$, $m_{g,r}$ is the average of $\operatorname{diag}\left(\mathbf{S}_{g, r}^{\mathrm{T}} \mathbf{S}_{g, r}\right)$, $v_{g,r}$ is the corresponding variance, and $\alpha_{g,r}$ is the significant level (here is 0.01) to derive the 99$\%$ confidence limit.

Besides the steady information, monitoring statistics $D_{g,s}^2$ and $D_{g,r}^2$ are defined for dynamic variations $\dot {\textbf S}_{g,s}$ and $\dot {\textbf S}_{g,r}$. Both of them can be defined as the traditional $SPE$ statistics with specific variances, which follow the $F$ distribution below,
\begin{equation}
\begin{aligned}
D_{g,s}^2 & =(\dot {\mathbf{S}}_{g,s})^{\mathrm{T}} \mathbf{\Lambda}_{g,s}^{-1}(\dot{\mathbf{S}}_{g,s}) \\
& \sim \frac{P_{g}(K_{g}^{2}-1)}{K_{g}(K_{g}-1)} F_{P_g, K_g-P_g, \alpha_{g,s}} \\
D_{g,r}^2 & =(\dot{\mathbf{S}}_{g, r})^{\mathrm{T}} \mathbf{\Lambda}_{g,r}^{-1} (\dot{\mathbf{S}}_{g,r}) \\
& \sim \frac{L_{g}(K_{g}^{2}-1)}{K_{g}(K_{g}-1)} F_{L_g, K_g-L_g,\alpha_{g,r}}
\end{aligned}
\end{equation}
where $\mathbf{\Lambda}_{g, s}^{-1}$ and $\mathbf{\Lambda}_{g, r}^{-1}$ are the inverse covariance matrixes of $\dot {\textbf S}_{g,s}$ and $\dot {\textbf S}_{g,r}$, respectively; $L_g$ is the number of latent variables in residual subspace of ${\widetilde {\textbf X}}_g$, which equals to $J-P_g$; $\alpha_{g,s}$ and $\alpha_{g,r}$ have the same meaning as that in Eqs. (12) and (13).

Eqs. (12) to (14) provide a deterministic diagnosis result to judge whether the sample is normal or not by checking the control limits. To consider the random nature of samples, the local probability indices inspired by Bayesian inference probability for a sample $\textbf x_k$ related to the $g^{th}$ Gaussian component are defined below,
\begin{equation}
\begin{array}{ll}
{p_{g,ss}^{local}(\mathbf{x}_{k})=p\{T_{g,s}^{2}(\mathbf{x}) \leq T_{g,s}^{2}(\mathbf{x}_{k})\}} \\ [1mm]
{p_{g,sr}^{local}(\mathbf{x}_{k})=p\{T_{g,r}^{2}(\mathbf{x}) \leq T_{g,r}^{2}(\mathbf{x}_{k})\}} \\ [1mm]
{p_{g,ds}^{local}(\mathbf{x}_{k})=p\{D_{g,s}^{2}(\mathbf{x}) \leq D_{g,s}^{2}(\mathbf{x}_{k})\}} \\[1mm]
{p_{g,dr}^{local}(\mathbf{x}_{k})=p\{D_{g,r}^{2}(\mathbf{x}) \leq D_{g,r}^{2}(\mathbf{x}_{k})\}}
\end{array}
\end{equation}
where $ss$ indicates the steady slow-varying subspace, $sr$ is the steady residual subspace, $ds$ is the dynamic slow-varying subspace, and $dr$ refers to the dynamic residual subspace.

\subsubsection{Development of the global probabilistic health indicator}
Given that a sample belongs to a particular pattern, local indices indicate the health state of the sample. However, this sample may come from multiple patterns due to the random nature. With the calculated posterior probabilities, four global indices are further defined to combine specific local probability metrics across all patterns below,
\begin{equation}
BIP_{*}=\sum_{g=1}^{G} \{p(\mathbf{x} \in g | \mathbf{x}_{k}) p_{g,*}^{\text {local}}(\mathbf{x}_{k})\}
\vspace{-0.1cm}
\end{equation}
where $*$ refers to any one of $ss$, $sr$, $ds$, and $dr$, respectively; $p(\mathbf{x} \in g | \mathbf{x}_{k})$ is the posterior probability that $\textbf x_k$ belongs to each operation pattern, which is calculated with the help of prior probability $\pi_g$ obtained in Eq. (4) through Bayesian inference strategy below,
\begin{equation}
p(\mathbf{x} \in g | \mathbf{x}_{k})=\frac{\pi_{g} p(\mathbf{x}_{k} | \boldsymbol{\mu}_{g}, \mathbf \Sigma_{g})}{\sum_{g=1}^{G} \{\pi_{g} p(\mathbf{x}_{k} | \boldsymbol{\mu}_{g}, \mathbf{\Sigma}_{g})\}}
\end{equation}

As $0\leq p_{g,_{*}}^{local}(\mathbf{x}_{k})\leq 1$ and $\sum_{g=1}^{G} p(\mathbf{x} \in g | \mathbf{x}_{k}) \leq 1$, we have,
\begin{equation}
0 \leq BIP_{*} \leq 1-\alpha
\end{equation}
where $\alpha$ is globally significant level, which is 0.01 here.

\subsection{Performance Evaluation}
If a new sample $\textbf x_{new}$ is incoming, its dynamic data vector $\textbf x_{d,new}$ can be constructed in the way given in Section III.A. After that, $\textbf x_{d,new}$ is projected into each pattern to calculate the corresponding steady and dynamic information. Taking the $g^{th}$ pattern as an example, its steady and dynamic information is computed as follows,
\begin{equation}
\begin{array}{l}
{\mathbf{s}_{new,g,s}=\mathbf{x}_{d,new}{\mathbf{W}_{g,s}}} \\ [1mm]
{\mathbf{s}_{new,g,r}=\mathbf{x}_{d,new}{\mathbf{W}_{g,r}}} \\ [1mm]
{\dot{\mathbf{s}}_{new,g,s}=\dot{\mathbf{x}}_{d,new}{\mathbf{W}_{g,s}}} \\ [1mm]
{\dot{\mathbf{s}}_{new,g,r}=\dot{\mathbf{x}}_{d,new}{\mathbf{W}_{g,r}}}
\end{array}
\end{equation}
where ${\dot {\mathbf x}}_{d,new}$ is the first order temporal difference of ${\mathbf x}_{d,new}$.

With Eq. (15), local probabilistic statistics at each pattern can be calculated accordingly, and they are integrated into global statistics with Eq. (16). Online process monitoring is sequentially conducted by continuously comparing all global statistics with their control limits, which are specified as 99$\%$ here. If all monitoring statistics stay within the normal region, this means that the IHP runs normally. Otherwise, if any statistics exceeds its control limit with three consecutive samplings, alarms will be released and the specific health status will be diagnosised. Health evaluation is carefully derived by analyzing the violated monitoring indices, corresponding to the evolution procedure from the normal status to the faulty status. Table II provides an explaination about how to combine monitoring indices to evaluate system health status. Further, the whole procedure of MSFA has been summarized in \textbf{Algorithm 1} for easy understanding.

Three quantitative indices widely used for evaluating the performance of process monitoring approaches are adopted here, which are false alarm rate ($FAR$), fault detection delay ($FDD$), and fault detection rate ($FDR$) [20]. $FAR$ evaluates the reliability of the developed model by checking its performance on normal testing data, and a low value is expected. Assuming that the number of samples in normal testing data is $N_N$, $FAR$ is calculated as $\frac{N_f}{N_N}$, where $N_f$ is the detected abnormal samples from $N_N$.

Both $FDD$ and $FDR$ measure the sensitivity of monitoring models to faulty data. $FDD$ informs the accuracy on the detection of the starting time of the fault. Assuming that the real fault occurrence time is $T_r$, $FDD$ is calculated as the error between $T_d$ of fault detection time ($FDT$) and $T_r$, i.e., $FDD=T_d - T_r$. A good $FDD$ should be close to real fault time but no early than it. $FDR$ reveals the ratio of detected abnormal samplings after the real fault is introduced. The higher the $FDR$ value the better. Denoting the total number of samples in fault data as $K_F$, $FDR$ is calculated as $\frac{K_f} {K_F}$, where $K_f$ is the detected abnormal samples from $K_F$.

{\color{blue}
\begin{algorithm}[!htb]
\caption{Mixture slow feature analysis}
\scriptsize
\LinesNumbered
\KwIn{Training time-series ${\mathbf X}$ and validation time-series $\mathbf X_v$}
\KwOut{The clustered continuous time-series with well estimated parameters $\hat {\mathbf \Theta}$, a global monitoring model by integrating slow feature models $\mathbf{W}_{g,s}$ and $\mathbf{W}_{g,r}$ at a proper value of $G$;}
Initialize $G$=2 \\
\For {G is less than $\sum_{m=0}^{M} C(M,m)$}
{
{\textbf{Part 1: Perform pattern clustering with temporal correlation and slowness (Section III.A)}}\\
\quad {Construct dynamic data matrix $\mathbf X_d$ and its first-order difference information $\dot {\mathbf X}_d$;} \\
\quad {Construct the comprehensive matrix $\mathbf X= [\mathbf X_d, \dot {\mathbf X}_d]$;} \\
\quad {Initialize parameters $\mathbf \Theta = \{ {\bm \mu_1, \bm \Sigma_1, \pi_1, \bm \mu_2, \bm \Sigma_2, \pi_2,..., \bm \mu_G, \bm \Sigma_G, \pi_G} \}$ used in GMM;} \\
\quad {Input $\mathbf X$ into GMM and update estimated parameters $\hat {\mathbf \Theta}$ using EM algorithm according to Eqs. (3) through (8);} \\
\quad {Assign samples of $\mathbf X$ into each cluster according to the poster probability and obtain subgroups $\{ \textbf X_1$, $\textbf X_2$, ..., $\textbf X_g$, ..., $\textbf X_G\}$;} \\
{\textbf{Part 2: A global monitoring model based on SFA (Section III.B)}}\\
\quad \For {$g = 1; g \le G$}
{
\quad {Calculate slow features with $\widetilde {\textbf X}_g$ and $\dot {\widetilde {\textbf X}}_g$ according to Eq. (10), where $\mathbf X_g =[\widetilde {\textbf X}_g, \dot {\widetilde {\textbf X}}_g]$;}\\
\quad {Denote the dominant slow features as ${\textbf S}_{g,s}={\widetilde {\textbf X}_{g,s}} {\mathbf W}_{g,s}$ and the remaining features as ${\textbf S}_{g,r}={\widetilde {\textbf X}_g} {\mathbf W}_{g,r}$;}\\
\quad {Calculate the corresponding slow-varying dynamic information as ${\dot{\mathbf{S}}_{g,s}=\dot {\widetilde {\textbf X}}_g \mathbf{W}_{g,s}}$ and ${\dot {\mathbf{S}}_{g,r}=\dot {\widetilde {\textbf X}}_g \mathbf{W}_{g,r}}$;}\\
\quad {Calculate control limits for two steady subspaces and two dynamic subspaces according to Eqs. (12) through (14), which are $T_{g,s}^{2}$, $T_{g,r}^{2}$, $D_{g,s}^{2}$, and $D_{g,r}^{2}$, respectively.}\\
}
\quad {Calculate the local probability of sampling $\mathbf x_k$ of a certain pattern as $p(\mathbf{x} \in g | \mathbf{x}_{k})=\frac{\pi_{g} p(\mathbf{x}_{k} | \boldsymbol{\mu}_{g}, \mathbf \Sigma_{g})}{\sum_{g=1}^{G} \{\pi_{g} p(\mathbf{x}_{k} | \boldsymbol{\mu}_{g}, \mathbf{\Sigma}_{g})\}}$;} \\
\quad {Calculate the global probability metrics across all patterns as $BIP_{*}=\sum_{g=1}^{G} \{p(\mathbf{x} \in g | \mathbf{x}_{k}) p_{g,*}^{\text {local}}(\mathbf{x}_{k})\}$, where $*$ refers to any one of $ss$, $sr$, $ds$, and $dr$, respectively;} \\
{\textbf{Part 3: Selection of a proper value for $G$ based on performance evaluation}}\\
\quad {Construct the comprehensive matrix for validation data, which is denoted as $\mathbf X_v$;} \\
\quad {Assign samples of $\mathbf V$ into the corresponding $G$ groups with the learned parameters $\hat {\mathbf \Theta}$ in GMM, and obtain subgroups $\{ \textbf V_1$, $\textbf V_2$, ..., $\textbf V_G\}$;} \\
\quad {Project $\{ \widetilde{\textbf V}_1$, $\widetilde{\textbf V}_2$, ..., $\widetilde{\textbf V}_G\}$} into the corresponding slow feature models $\mathbf{W}_{g,s}$ and $\mathbf{W}_{g,r}$ calculated in Part 2; \\
\quad {Calculate local monitoring statistics and integrate them into a series of global monitoring statistics $BIP_{ss}$, $BIP_{sr}$, $BIP_{ds}$, and $BIP_{dr}$;}\\
\If {$BIP_{ss}$, $BIP_{sr}$, $BIP_{ds}$ and $BIP_{dr}$ are all below their control limits (0.99 here) for all samplings}
{The value $G$ is proper and stop iteration;}
\If {Anyone of $BIP_{ss}$, $BIP_{sr}$, $BIP_{ds}$ and $BIP_{dr}$ beyond its control limit (0.99 here) in continuous three samplings}
{Increase $G$ by one and go the Part 1 for the next iteration;}
}
\textbf{final}
\end{algorithm}
}

\section{Experimental Setup and Results}
In this section, the proposed MSFA has been implemented on a practical IHP for daily hot water supply. We first verify the efficacy of sequential 
clustering, followed by the model evaluation at the normal status with new patterns. When the system evolves into failure stage, fault detection ability of MSFA has been compared with the mainstream approaches.

\subsection{Data Description and Parameter Configuration}
According to the experimental setup given in Section II, four months of data are collected from April 2018 to July 2018 for analysis. With the sampling interval of 1 minute, 1440 samples of inlet water temperature and outlet water temperature are collected in a day. The data on April 1 through 20 are employed as training data to develop the corresponding monitoring models, and the validation dataset consists of the measurements from April 21 through 30 to determine crucial parameters. The remaining data collected from the other three months are used for model updating and further testing.

\begin{table}[!ht]
\renewcommand{\arraystretch}{1.2}
\setlength{\abovecaptionskip}{-0.5cm}
\setlength{\belowcaptionskip}{-0.8cm}
\scriptsize
\begin{center}
\caption{The crucial tunable parameters defined in the proposed method and ways for tuning them.}
\end{center}
\label{table_V}
\begin{center}
	\begin{threeparttable}
		\begin{tabular} {p{1cm} p{2cm} p{4cm} p{8cm}}
		\toprule
		{\textbf{Symbol}} & {\textbf {Location}} & {\textbf{Description}} & {\textbf{Guidance for tunning}}\\
		\hline
		{$h$} & {Section III.A} & {Time lag} & {Using the root summed squares of all variables from the normal data against a certain confidence bound ($\pm 1\%$ here)}\\
                \hline
		{$G$} & {Section III.A} & {The number of Gaussian component} & {Using a step-wise way to iteratively find the proper value by checking the monitoring performance using validation data and details are given in \textbf{Algorithm 1}}\\
                \hline
		{$P_g$} & {Section III.B} & {The number of retained slow features in each Gaussian model} & {Finding the “knee" point regarding slowness from the singular value $\lambda$ calculated from Eq. (10) in each Gaussian model}\\
		\bottomrule
		\end{tabular}
	\end{threeparttable}
\end{center}
\vspace{-0.5cm}
\end{table}

\begin{table}[!ht]
\renewcommand{\arraystretch}{1.2}
\setlength{\abovecaptionskip}{-0.8cm}
\scriptsize
\begin{center}
\caption{The specific values of crucial tunable parameters used in the proposed method and its counterparts.}
\end{center}
\label{table_V}
\begin{center}
	\begin{threeparttable}
		\begin{tabular} {c c c}
		\toprule
		{\textbf{Method}} & {\textbf{Symbol or name}} & {\textbf {Value}}\\
		\hline
		\multirow{3}{*}{MSFA (Proposed)} & {$h$} & {43} \\
			& {$G$} & 6 \\
			& {$P_g$} & 10, 5, 5, 5, 5, 5 \\
		\hline
		GMM & The number of Gaussian component & 8\\
		\hline
		\multirow{2}{*}{DPCA} & Time lag & 43\\		
			& Retained PCs in system subspace & 6 \\
		\hline			
		\multirow{2}{*}{CVA} & Time lag & 43\\	
			& Retained latent variables in system subspace & 6 \\
		\hline
		\multirow{2}{*}{SFA} & Time lag & 43\\
			& Retained slow features in system subspace & 6 \\		
		\hline
		\end{tabular}
	\end{threeparttable}
\end{center}
\vspace{-0.4cm}
\end{table}

The crucial parameters defined in the proposed method are summarized in Table III, in which ways for tuning them are also provided. Specifically, by adopting the sample auto-correlation function, time lags $h$ of the two temperature measurements are both determined as 43 using training data. On the basis of this, the dynamic data matrix is constructed by lagging original measurements, and then integrating temporal difference information, generating a data matrix with the variable dimension of 172. According to the different combinations of the five ASHP units, the total number of possible physical operation patterns is 32. According to the given steps in {\textbf{Algorithm 1}}, similar Gaussian components will be merged with the assist of monitoring performance, six Gaussian components are finally retained, each of which corresponds to a specific varying trajectory. The specific value of retained slow features in each pattern is calculated as 10, 5, 5, 5, 5, and 5 using the validation data, respectively.

All offline model development and online data analysis are conducted on a workstation equipped with 16 processors of Intel Xeon E5-2620 v4 (20 MB cache, up to 2.10 GHz) and a multi-graphics processor unit of NVIDIA GeForce GTX 1080Ti (11 GB). The programming software is Matlab with the version of 2019a.

\begin{figure}[htb]
\subfigure[]
{
\begin{minipage}[t]{0.4\linewidth}
\centering
\includegraphics[width=6cm]{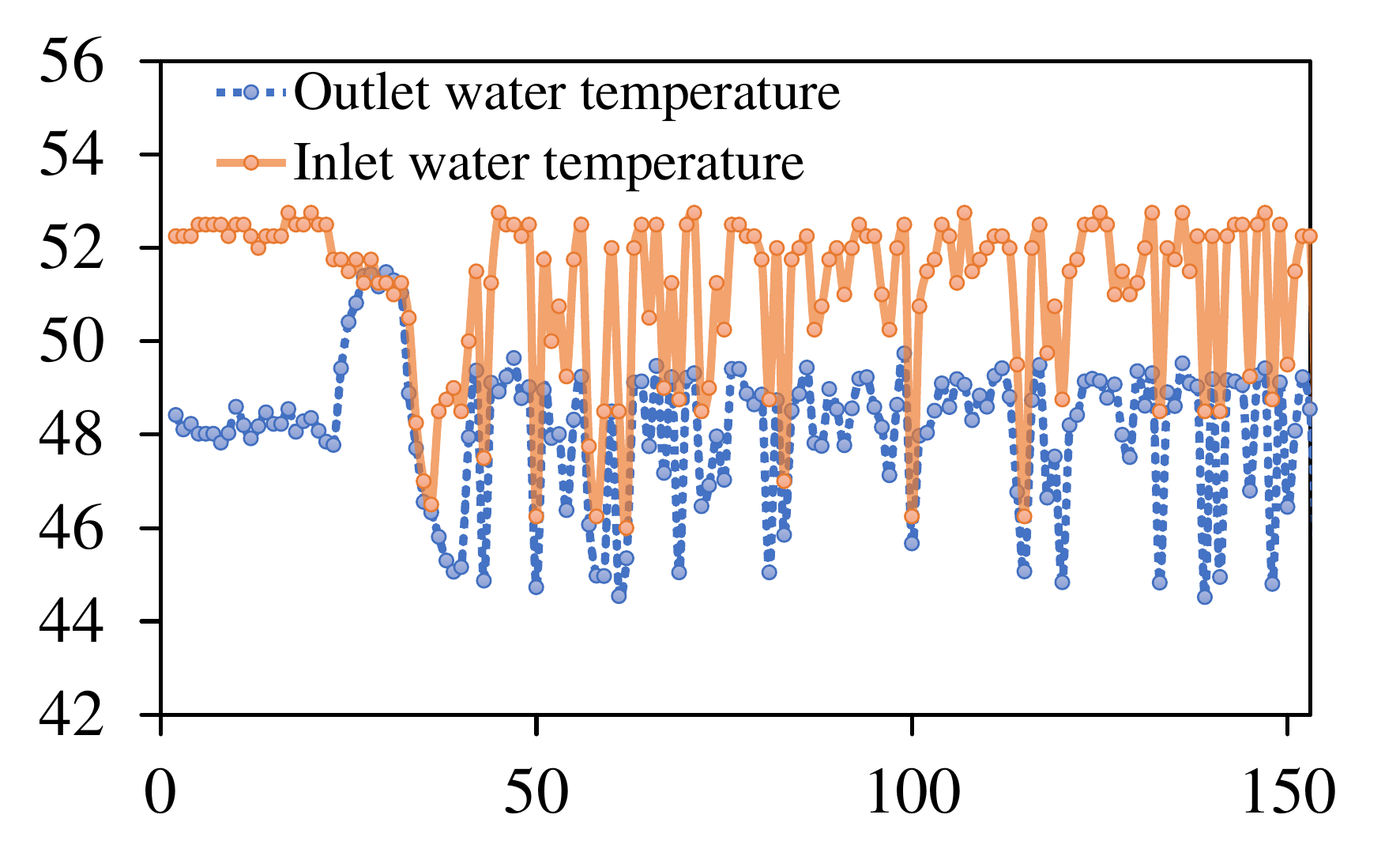}
\end{minipage}
}
\subfigure[]
{
\begin{minipage}[t]{0.4\linewidth}
\centering
\includegraphics[width=6cm]{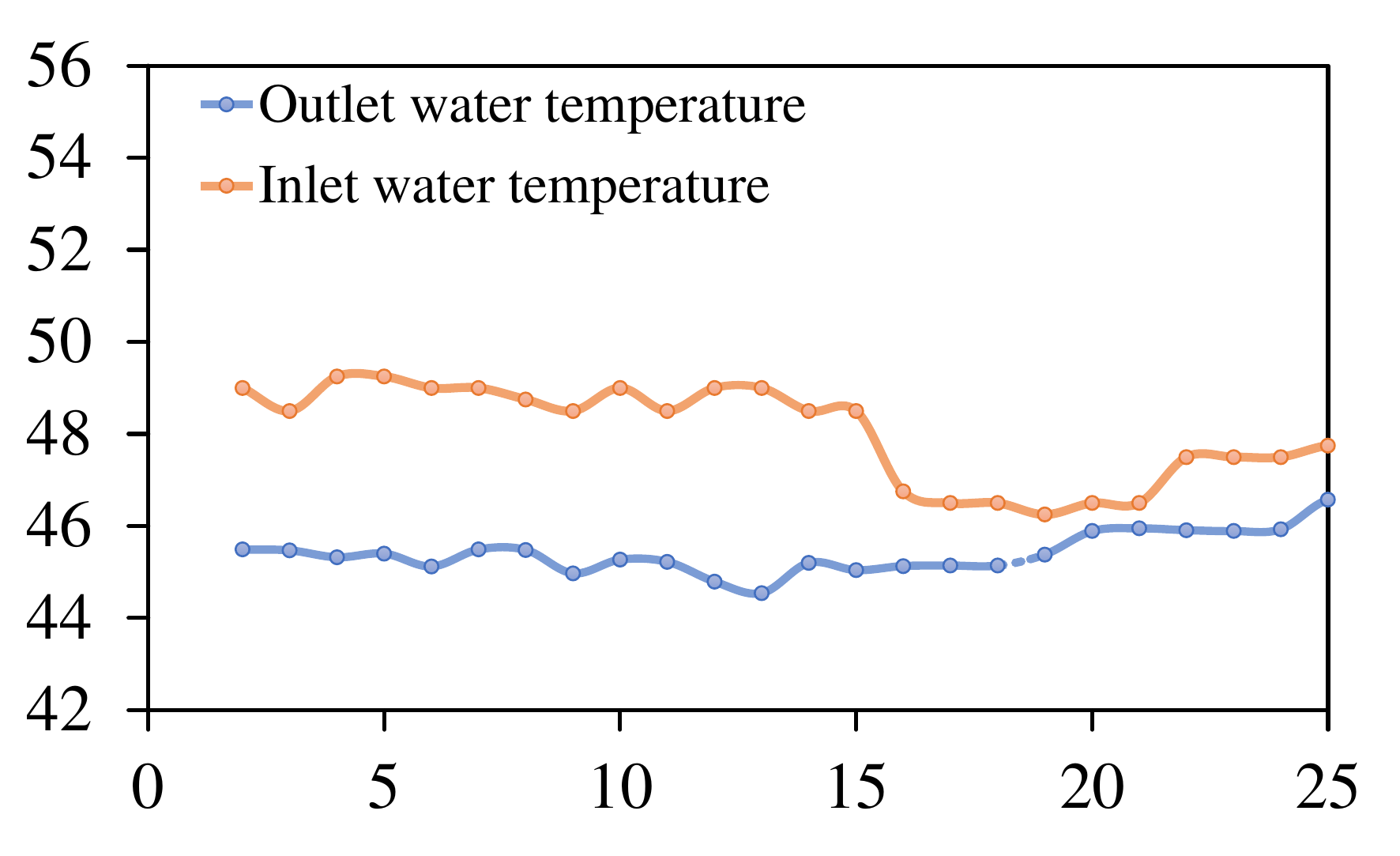}
\end{minipage}
}
\hfill
\subfigure[]
{
\begin{minipage}[t]{0.4\linewidth}
\centering
\includegraphics[width=6cm]{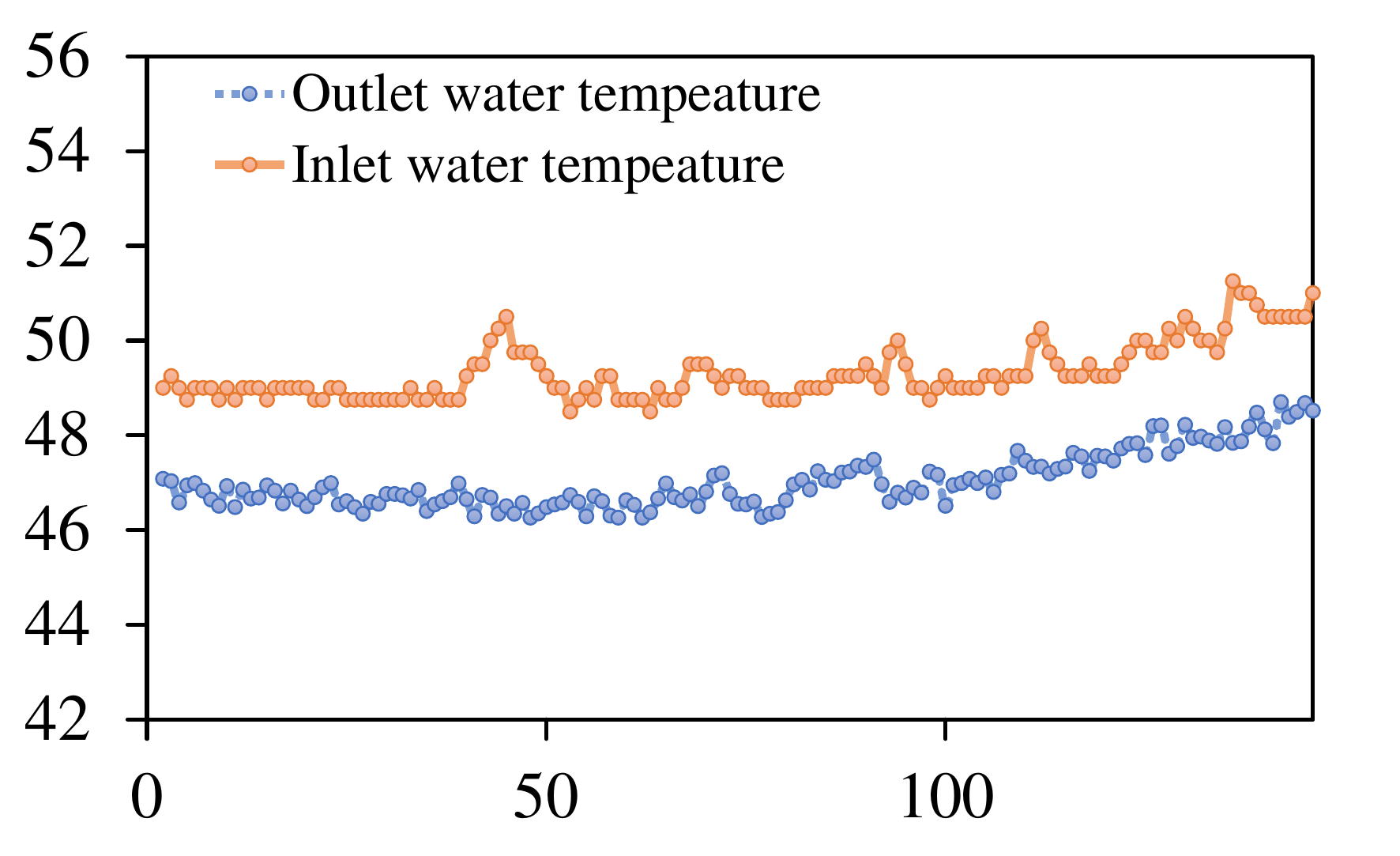}
\end{minipage}
}
\subfigure[]
{
\begin{minipage}[t]{0.4\linewidth}
\centering
\includegraphics[width=6cm]{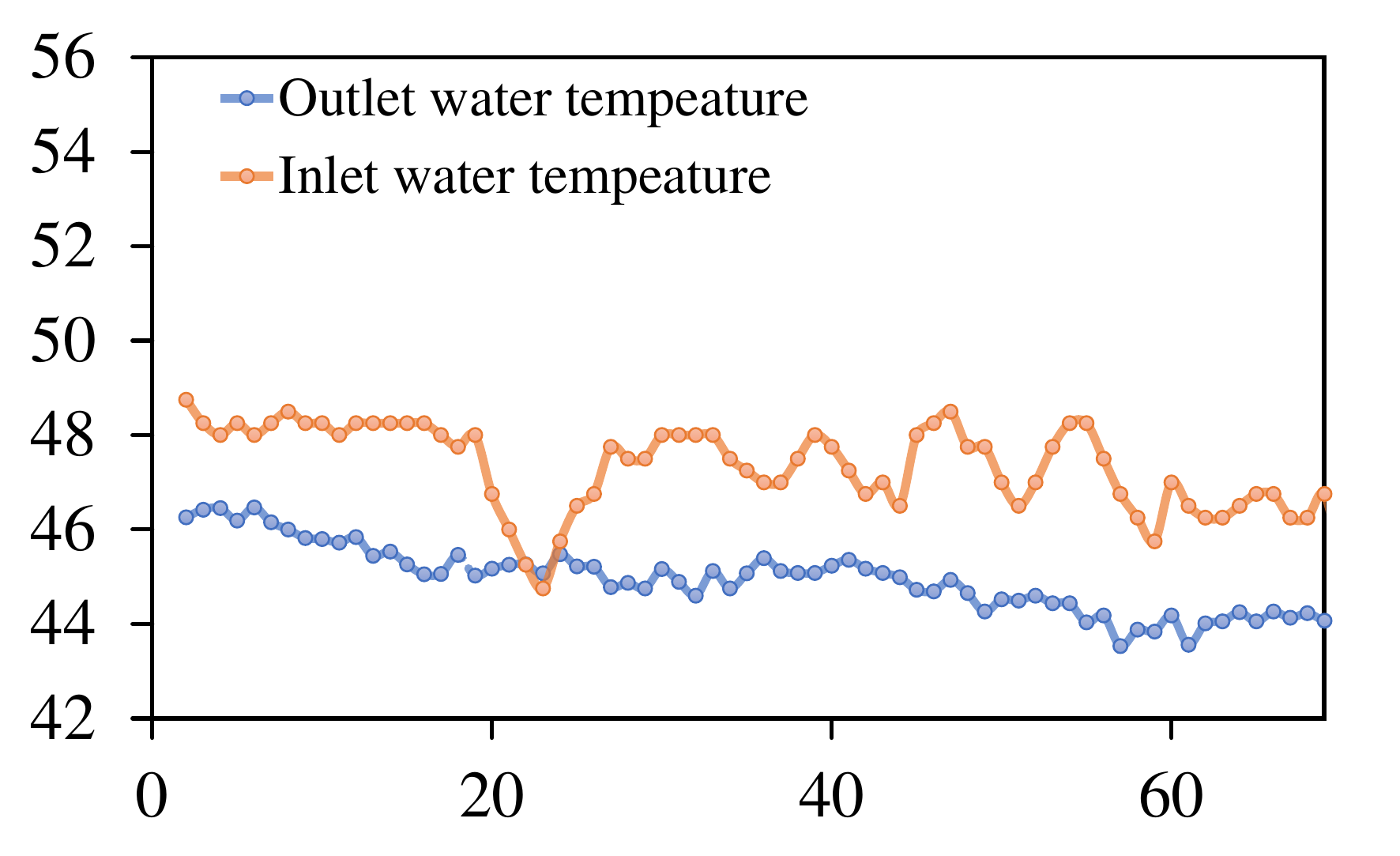}
\end{minipage}
}
\hfill
\subfigure[]
{
\begin{minipage}[t]{0.4\linewidth}
\centering
\includegraphics[width=6cm]{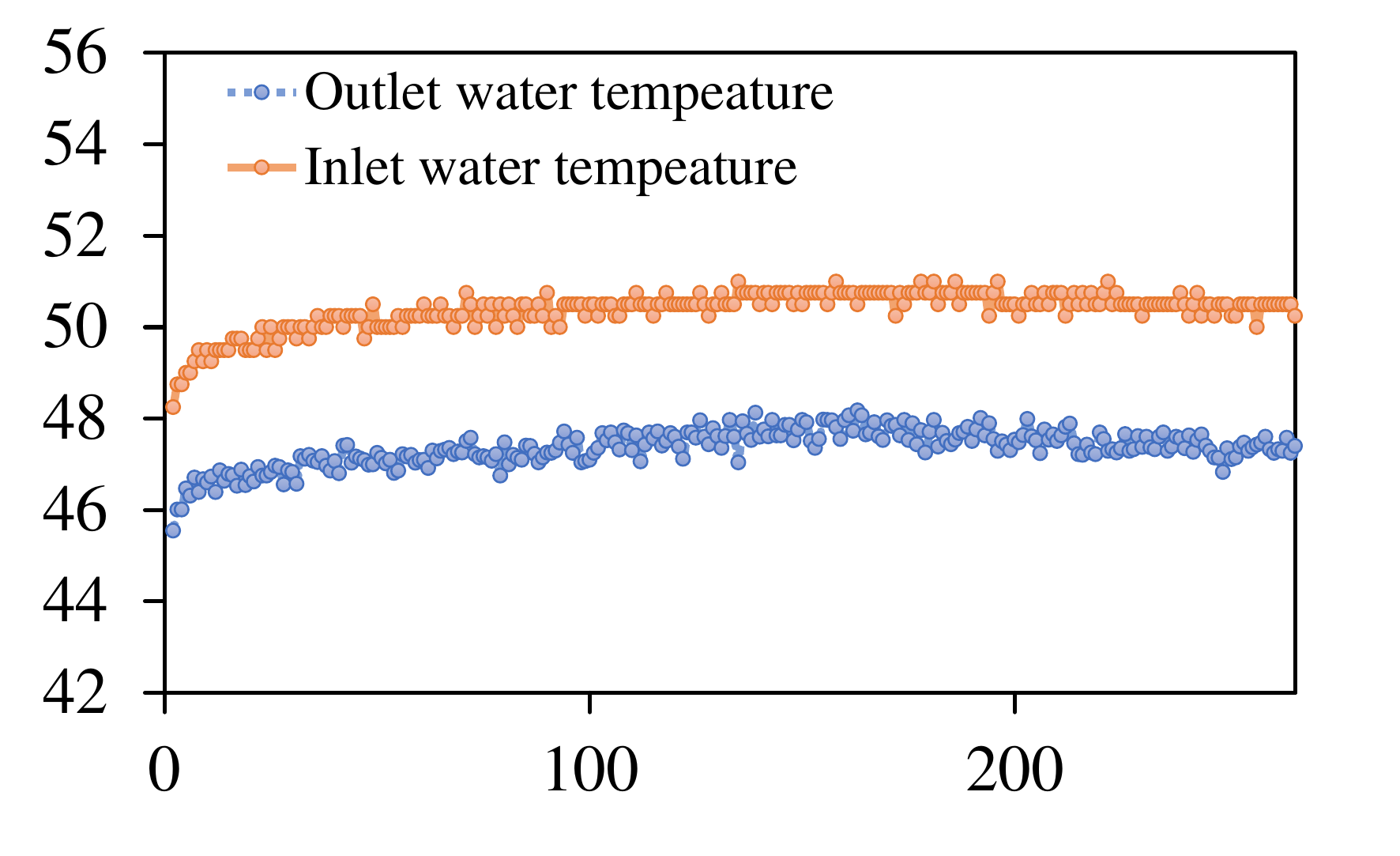}
\end{minipage}
}
\subfigure[]
{
\begin{minipage}[t]{0.4\linewidth}
\centering
\includegraphics[width=6cm]{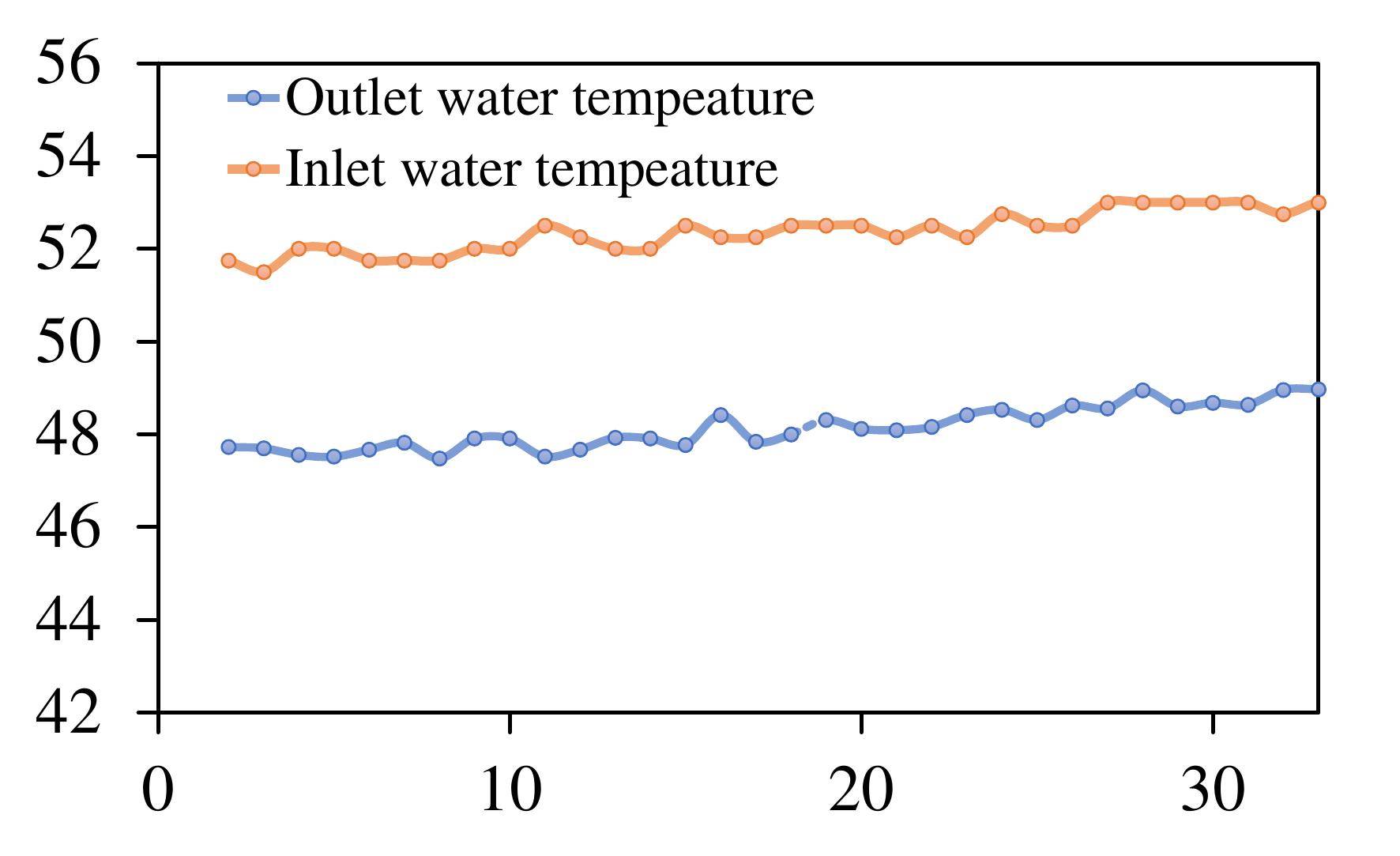}
\end{minipage}
}
\centering
\caption{Varying trend of water temperature in classified patterns for (a) Pattern 1, (b) Pattern 2, (c) Pattern 3, (d) Pattern 4, (e) Pattern 5, and (6) Pattern 6.}
\vspace{-0.3cm}
\end{figure}

\begin{figure}[!htb]
\subfigure[]
{
\begin{minipage}[t]{1\linewidth}
\centering
\includegraphics[width=8cm]{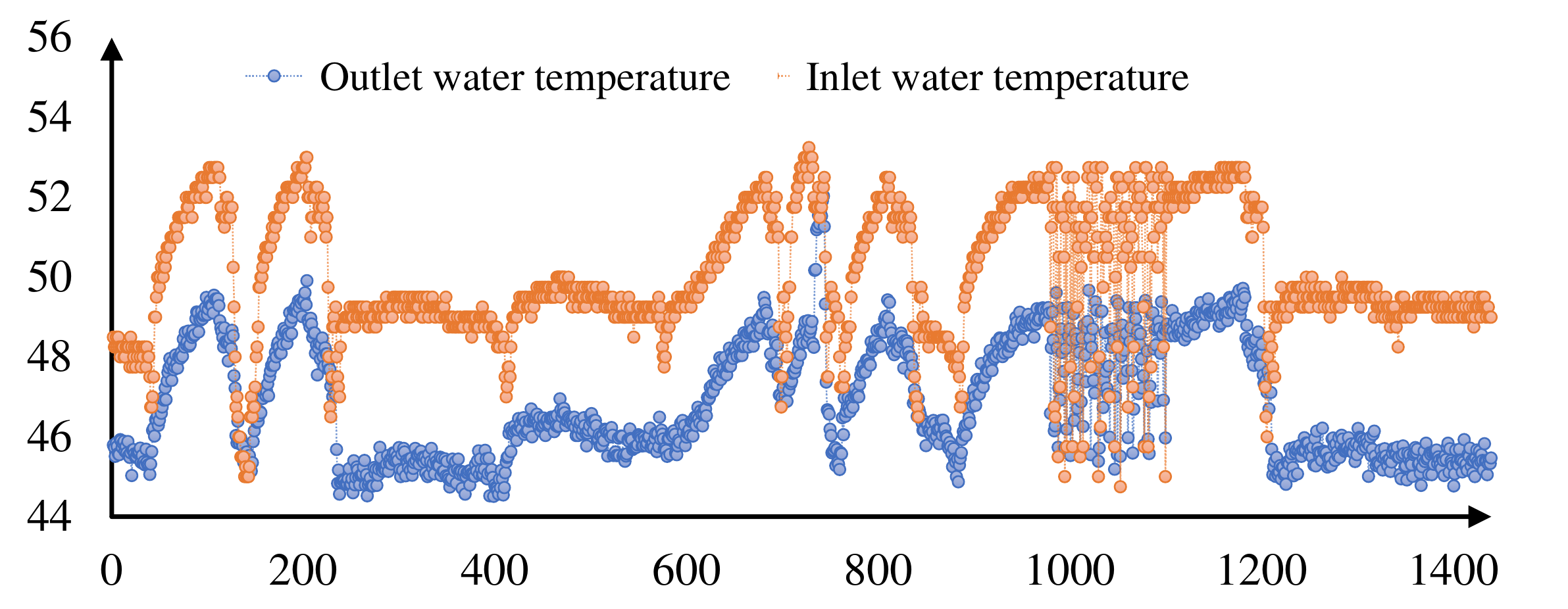}
\end{minipage}
}
\hfill
\subfigure[]
{
\begin{minipage}[t]{1\linewidth}
\centering
\includegraphics[width=8cm]{figure/FIG5a_TII-20-5401.pdf}
\end{minipage}
}
\hfill
\subfigure[]
{
\begin{minipage}[t]{1\linewidth}
\centering
\includegraphics[width=8cm]{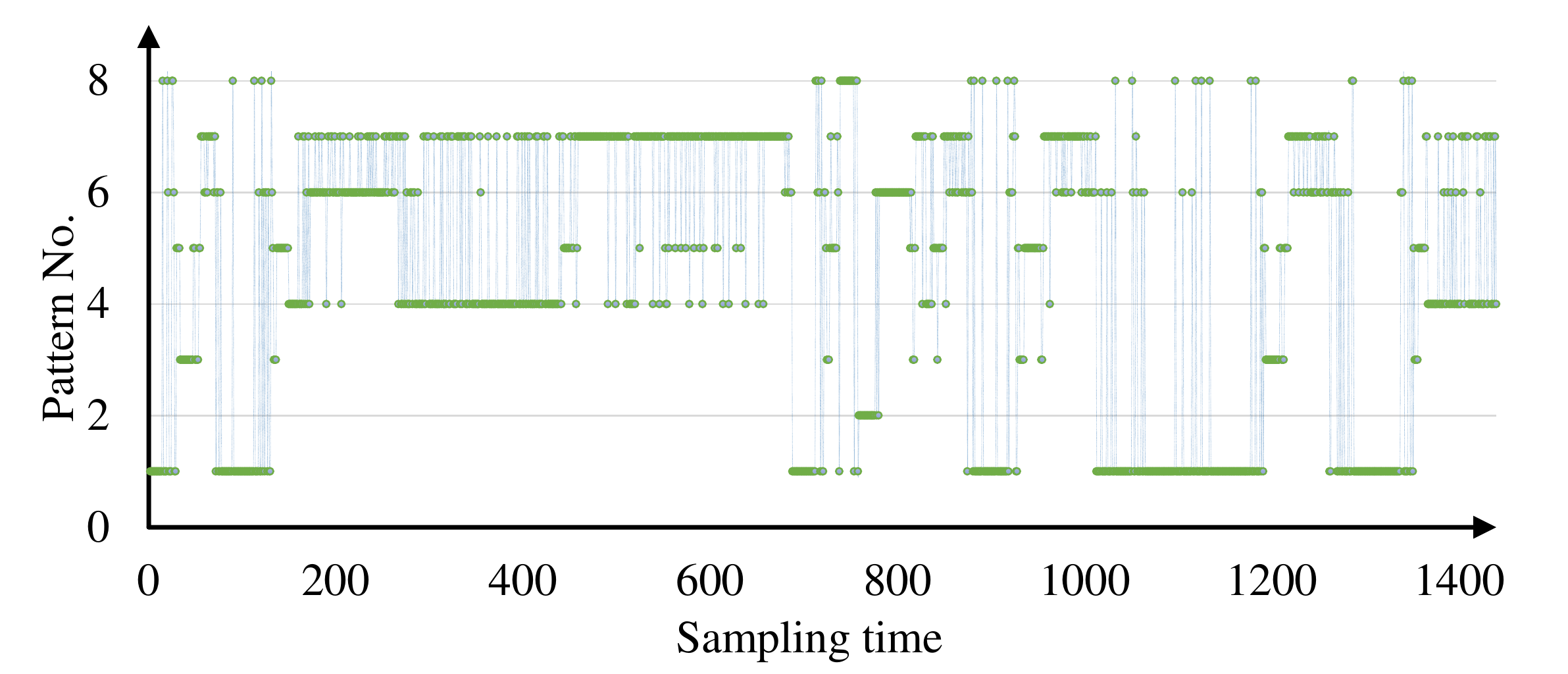}
\end{minipage}
}
\centering
\caption{Illustration of (a) measurements collected on 21 April, (b) the sequential clustering results using the proposed method on 21 April data, and (c) the clustering results of 21 April data using GMM.}
\vspace{-0.6cm}
\end{figure}

\begin{figure}[htb]
\subfigure[]
{
\begin{minipage}[t]{1\linewidth}
\centering
\includegraphics[width=8cm]{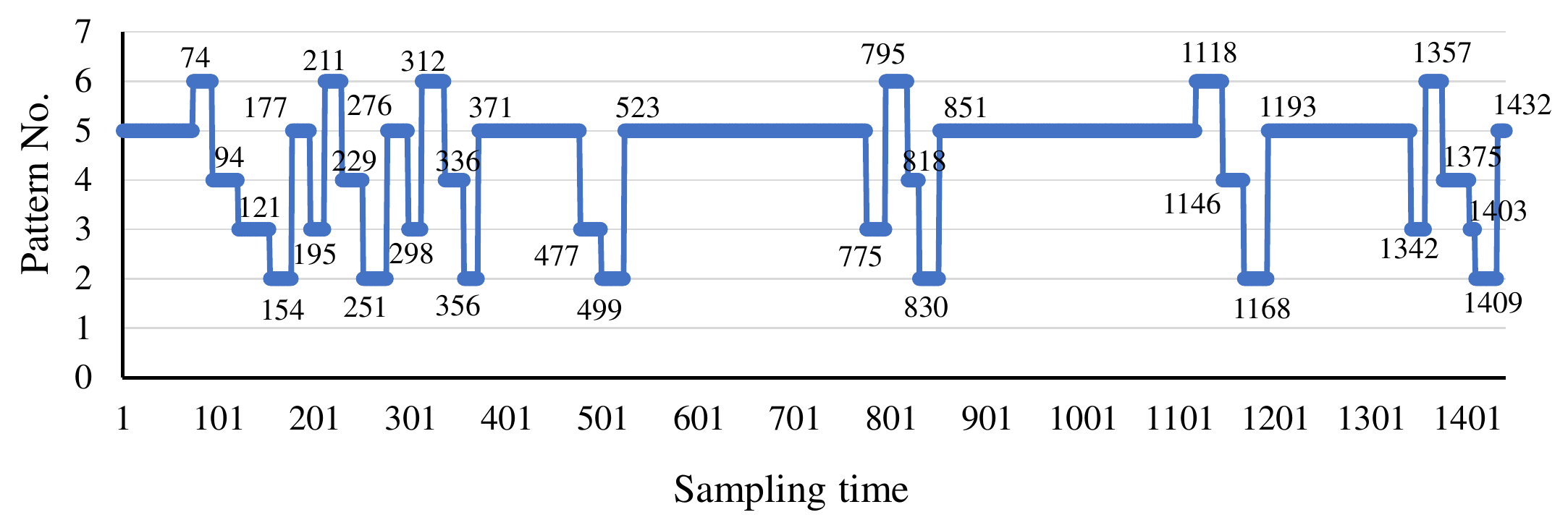}
\end{minipage}
}
\hfill
\subfigure[]
{
\begin{minipage}[t]{1\linewidth}
\centering
\includegraphics[width=8cm]{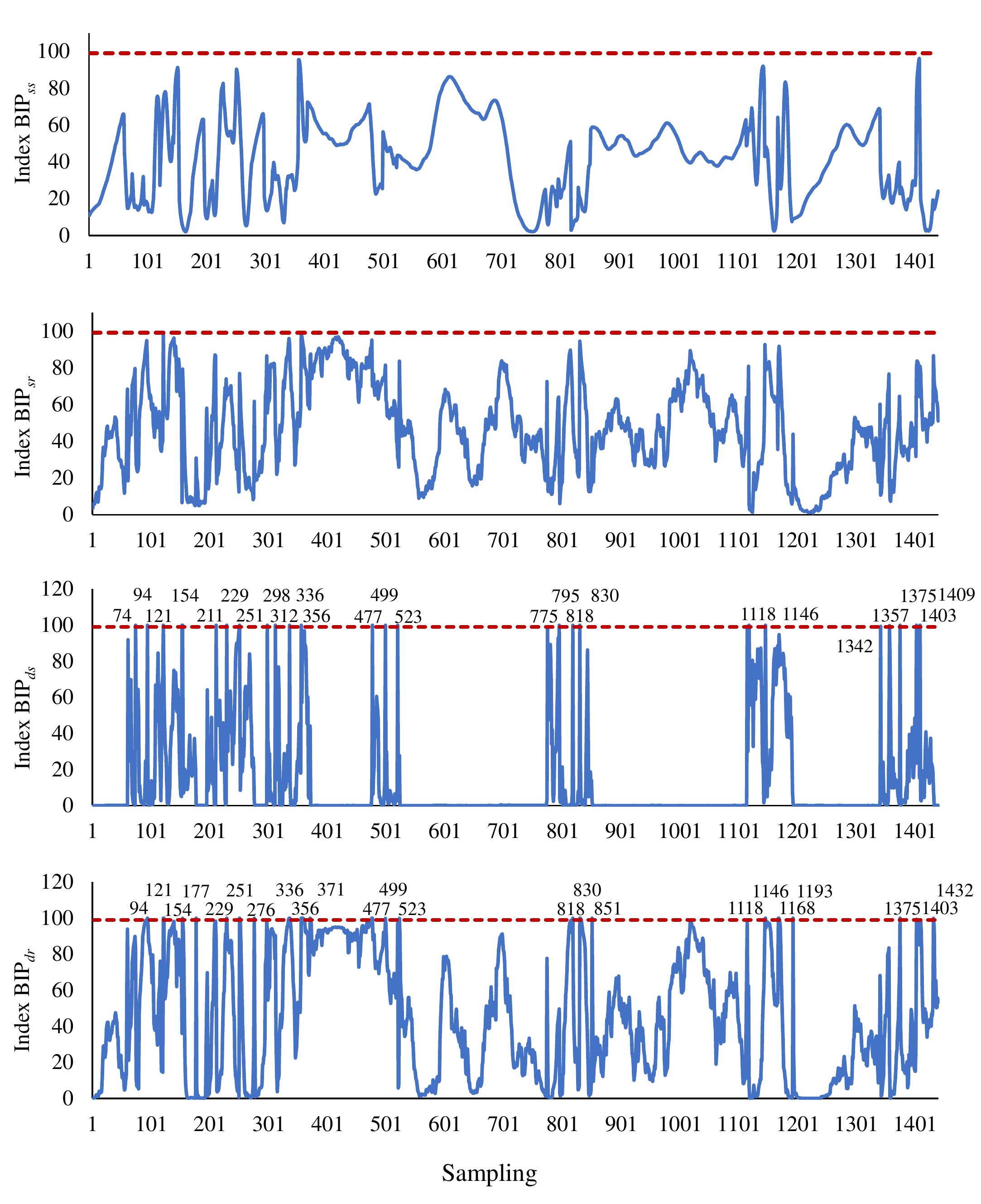}
\end{minipage}
}
\caption{Performance of the proposed method on April 30 data concerning (a) sequential clustering results and (b) monitoring results with defined indices.}
\vspace{-0.7cm}
\end{figure}

\subsection{Sequential Pattern Classification}
Fig. 4 depicts an example of the clustered six patterns, each of which presents unique varying trend. Specifically, in Pattern 1, the outlet water temperature rapidly fluctuates between 46$^\circ$C and 52$^\circ$C. In Pattern 2, the water temperature decreases slowly and reaches the minimum temperature at the end of the procedure. For Patterns 3 and 4, the water temperature presents a fast increasing and declining trajectory, respectively. The varying trend of the water temperature in Pattern 5 is relatively slow and almost stable at the 50.5$^\circ$C. The water temperature increases slowly to the maximum in Pattern 6. Similar conclusions can be drawn for the inlet water.

\begin{table*}[!ht]
\renewcommand{\arraystretch}{1.2}
\setlength{\abovecaptionskip}{-0.1cm}
\setlength{\belowcaptionskip}{-0.2cm}
\centering
\scriptsize
\caption{Monitoring performance for the normal testing data with respect to False Alarm Rate (FAR) $\%$.}
\begin{tabular}{c| c c| c c c c| c c| c c c c}
\hline
\multicolumn{1}{c|}{\multirow{2}{*}{\textbf{Time}}} & \multicolumn{2}{c|}{\textbf{DPCA}} & \multicolumn{4}{c|}{\textbf{SFA}} & \multicolumn{2}{c|}{\textbf{CVA}} & \multicolumn{4}{c}{\textbf{MSFA (Proposed)}}\\
\cline{2-13}
\textbf{} & $T^2$ & $SPE$ & $T_d^2$ & $T_e^2$ & $S_d^2$ & $S_e^2$ & $T^2$ & $SPE$ & $T_d^2$ & $T_e^2$ & $S_d^2$ & $S_e^2$ \\
\hline
{From May 1 to May 31} & 7.27 & 0.14 & 1.02 & 0.13 & 0.00 & $6.56$ & 0.00 & 0.00 & 0.37 & 0.11 & 0.58 & 0.79 \\
\hline
{From June 1 to June 28} & 8.37 & 0.00 & 8.70 & 0.00 & 0.00 & 3.08 & 0.00 & 0.00 & 0.62 & 0.41 & 0.53 & 0.53 \\
\hline
\end{tabular}
\vspace{-0.4cm}
\end{table*}

With the well-trained sequential clustering model, measurements on 21 April, as shown in Fig. 5(a) are used for sequential clustering. Frequent switching is observed among the patterns, as shown in the results given in Fig. 5(b). And more discoveries could be revealed after careful analysis. First, presenting a stable behavior, Pattern 5 appears most, which corresponds to the temperature of the set-point. Second, Pattern 4 is always followed by Pattern 2, indicating that the minimum water temperature is achieved after temperature continually drops. Third, Pattern 3 is always followed by Pattern 6 or Pattern 5, which means the higher water temperature is observed after increasingly heating. Using the same training dataset and validation dataset, the original measurements are employed as input of GMM for comparison. As shown in Fig. 5(c), the clustering results of GMM are discontinuous.

\subsection{Health Monitoring using the Proposed MSFA}
\subsubsection{Identification of pattern switching}
With classified patterns shown in Fig. 6(a), health evaluation results on 30 April data are given, as shown in Fig. 6(b). Indices $BIP_{ss}$ and $BIP_{sr}$ reflect the monitoring statistics of steady subspace, both of which are below their control limits, indicating the normal status of steady distribution. Indices $BIP_{ds}$ and $BIP_{dr}$ measure the varying speed of the slow-varying and fast-varying process variations. It is observed that values of $BIP_{ds}$ and $BIP_{dr}$ are occasionally over their control limits at certain sampling times, which have been specified in Fig. 6(b). According to the judgment rules given in {Health status I} in Table II, the instantaneous abnormality of varying dynamics is caused by switching among the existing patterns. By comparing the results of classified patterns, these labeled samples are just the switch points between adjacent patterns, indicating the accurate classification ability of MSFA.

\begin{figure}[htb]
\subfigure[]
{
\begin{minipage}[t]{1\linewidth}
\centering
\includegraphics[width=8cm]{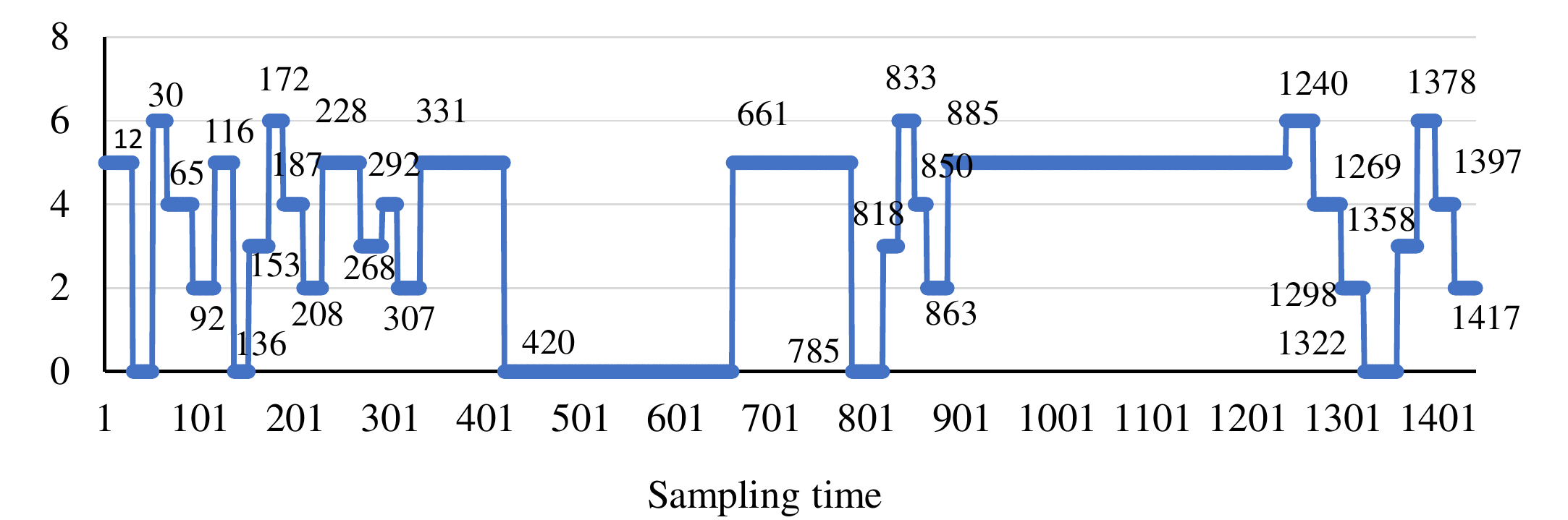}
\end{minipage}
}
\hfill
\subfigure[]
{
\begin{minipage}[t]{1\linewidth}
\centering
\includegraphics[width=8cm]{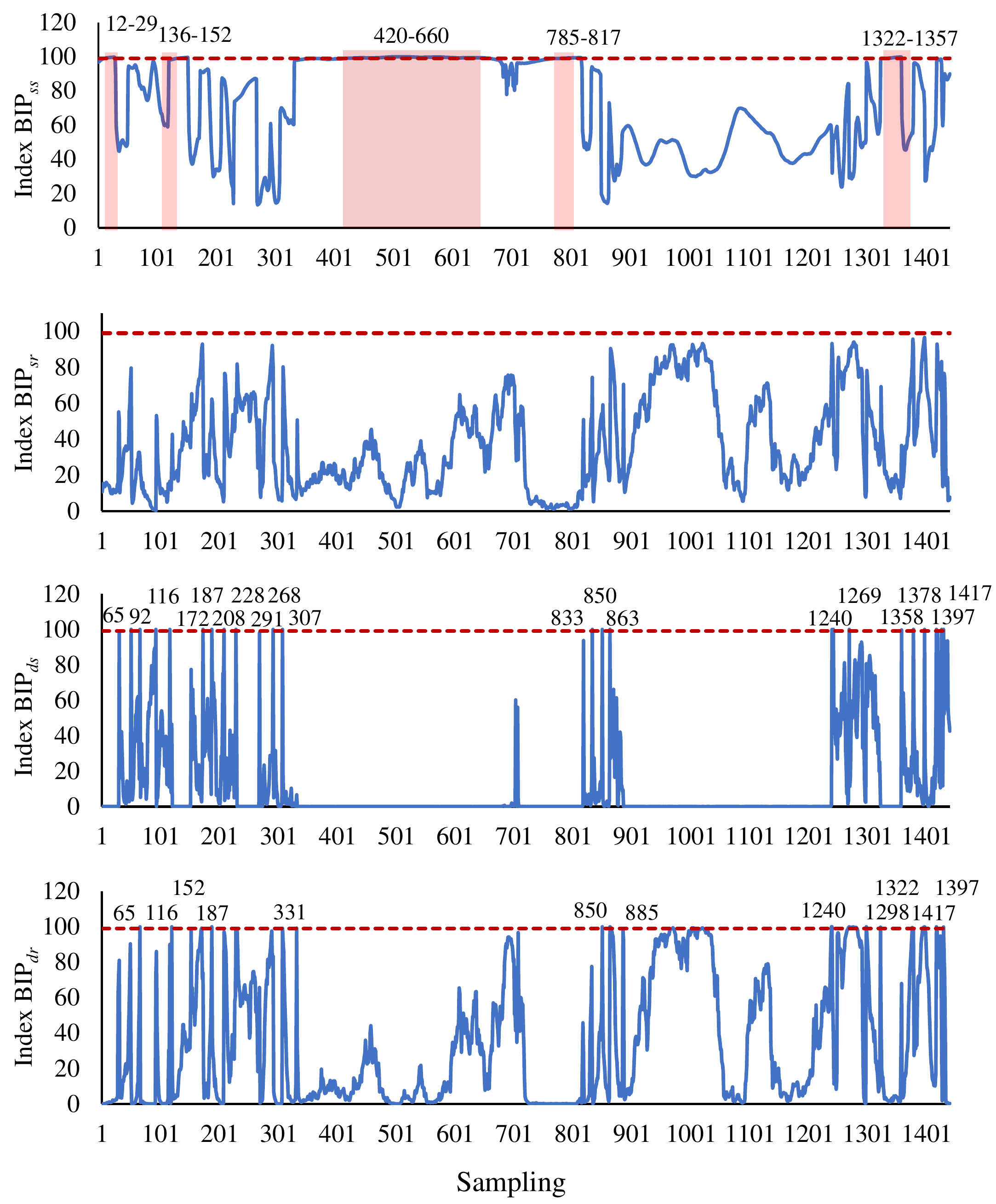}
\end{minipage}
}
\subfigure[]
{
\begin{minipage}[t]{1\linewidth}
\centering
\includegraphics[width=8cm]{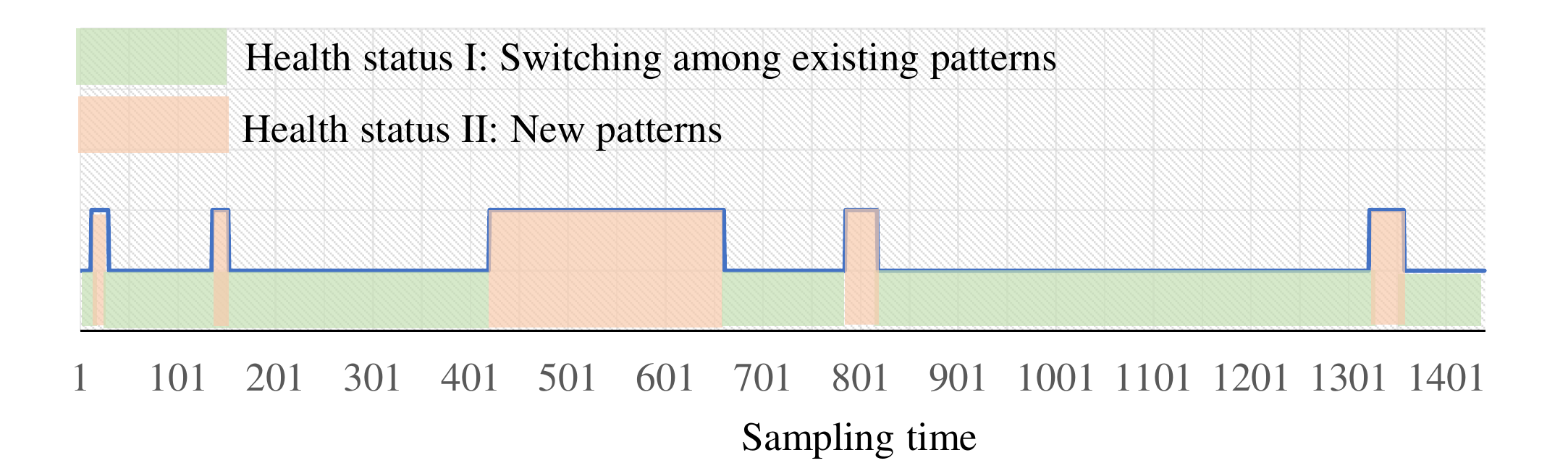}
\end{minipage}
}
\caption{Performance of the proposed method on May 23 data concerning (a) sequential clustering results, (b) monitoring results with defined indices, and (c) health evaluation results.}
\vspace{-0.7cm}
\end{figure}

\subsubsection{Model updating with new patterns}

We apply the proposed MSFA on data collected in May and June to demonstrate the strengths of the proposed method in comparison with typical algorithms, dynamic PCA (DPCA) \cite{Ref15}, canonical variate analysis (CVA) \cite{Ref17}, SFA \cite{Ref18}. Besides, the specific values of parameters used in these methods are given in Table IV. The IHP system normally runs during the first 22 days in May except the pattern switching. However, continuous alarms have been reported on 23 May, and specifics can be found in Fig. 7. Although monitoring statistics of $BIP_{sr}$, $BIP_{ds}$, and $BIP_{dr}$ are within their normal region, index $BIP_{ss}$ becomes abnormal at the 12$^{th}$ sample for the first time, and then this abnormality is repeated several times. According to the judgment rules given in {Health status II}, a new pattern appears since no dynamic abnormalities are observed. By including the new pattern into the existing patterns, MSFA is renewed, and the monitoring results on 24 May become normal again, where monitoring statistics of all indices become normal.

\begin{table}[!ht]
\renewcommand{\arraystretch}{1.2}
\setlength{\belowcaptionskip}{-0.2cm}
\centering
\scriptsize
\caption{Monitoring performance for the faulty data with respect to Fault Detection Time ($FDT$) and Fault Detection Rate ($FDR$).}
\begin{tabular}{c c c c c}
\hline
\textbf{Method} & \textbf{Statistics} & {\textbf{FDT}} & {\textbf{FDR ($\%$)}} \\
\hline
\multirow{2}*{\textbf{DPCA}} & $T^2$ & 22h2min on July 28 &17.81 \\
& $SPE$ & 21h27min on July 28 & 1.92 \\
\hline
\multirow{4}*{\textbf{SFA}} & $T_d^2$ & 22h35min on July 1 & 31.09 \\
& $T_e^2$ & 22h1min on July 28 & 1.89 & \\
& $S_d^2$ & No alarm detected & 0  & \\
& $S_e^2$ & 23h13min on July 2 & 5.00 & \\
\hline
\multirow{2}*{\textbf{CVA}} & $T^2$ & 22h42min on July 28 & 0.09 \\
& $SPE$ & 22h30min on July 28 & 1.85 \\
\hline
\multirow{4}*{\textbf{Proposed}} & $BIP_{ss}$ & 2h11min on June 29 & 60.96 \\
& $BIP_{sr}$  & 2h10min on July 3 & 2.54 \\
& $BIP_{ds}$ & 0h38min on July 1 & 3.94  \\
& $BIP_{dr}$ & 0h50min on June 30 & 7.22 \\
\hline
\end{tabular}
\vspace{-0.6cm}
\end{table}

New patterns will be updated into the existing patterns once detected. Collecting samples in a new pattern within one day, the local monitoring model of this new pattern could be derived using Eqs. from (9) to (11), and corresponding control limits are calculated with Eq. (15). Finally, the global probabilistic health indicators are updated using Eq. (16) with learned weights. Due to the limitation of pages, details of pattern updating are not given here, but it is an important issue that deserves more attention to develop intelligent ways, such as recursive manner.

Table V summarizes the results in terms of index $FAR$ for normal testing data in May and June. A lower $FAR$ indicates better offline modeling. Moreover, a large value of $FAR$ will weaken the reliability of the developed monitoring model. Especially if the value of $FAR$ beyond the significant level of control limits, it means that the monitoring results of the developed model are not reliable. Since the $FAR$ of DPCA is larger than the significant level (1$\%$), the developed DPCA model is not reliable. And similar conclusion could be drawn for SFA in June. Although the value of $FAR$ for CVA is low, this attributes to the insensitive monitoring ability.

\subsubsection{Performance evaluation with faulty case}
According to the maintenance records, the fourth ASHP broke down totally at the end of July and remained shutdown ever since. Besides, the third and fifth ASHPs got some problems at the end of July for further repair \cite{Ref12}. Since it is breakdown maintenance in practice, accurate fault occurrence time is unavailable. Nevertheless, it is reasonable to infer that a long degradation behavior may exist before ASHPs thoroughly out of work.

Reliable abnormality is observed for the first time on 29 June by MSFA. The faulty status at this time is not very serious since the dynamic index $BIP_{dr}$ returns to the normal region quickly. It is possible that only partial ASHPs got problems, leading to limited patterns are influenced. With the increasing deterioration of health status, more severe fault status is observed at the end of July, corresponding to a long period abnormal alarms. Especially, steady index $BIP_{ss}$ and dynamic index $BIP_{dr}$ both are above the control limits, indicating faulty status according to the judgment rule given in {Health status IV} in Table II. 

Although the monitoring results of SFA and DPCA are no longer reliable, we still compare their $FDT$ in Table VI for fairness. Both methods release fault signals at the end of July when the fault status is severe and the ASHP is nearly broken down. For CVA, due to its efficacy in dealing with dynamic time-series, its $FAR$ is almost zero, as shown in Table V. However, it is insensitive to the faulty data, and the first alarm is released on July 28. Therefore, the monitoring results of the proposed MSFA not only meet well with the actual case but also could identify the fault at a very early stage.

\section{Conclusion}
In this paper, the health degradation problem of integrated heat pumps was approached from the comprehensive perspectives of multi-mode and inherently dynamic slowness characteristics, and a mixture slow feature analysis method was proposed. The analysis of multi-pattern property and slowness contribute to investigate and separate the multiple varying trends, thereby avoiding mixing multiple time series correlations and enabling elaborated modeling in each pattern. A Gaussian mixture model-based solution algorithm was proposed to tackle the clustering problem. Further, the comprehensive monitoring from the steady and dynamic perspective is achieved by performing slow feature analysis in each mode. A global monitoring model is gained by integrating each local monitoring model. In a practical scenario, experiment studies demonstrated that the proposed method yields fine-grained clustering results and provides timely health status before the equipment breaks down.

\ifCLASSOPTIONcaptionsoff
  \newpage
\fi

\end{document}